\documentclass{article}

\usepackage{microtype}
\usepackage{graphicx}
\usepackage{subfigure}
\usepackage{booktabs} 
\usepackage[frozencache]{minted}
\usepackage{lscape}
\usepackage{amsmath}
\usepackage{amssymb}
\usepackage{natbib}

\usepackage[hyphens]{url}
\usepackage[hidelinks]{hyperref}

\usepackage[accepted]{icml2021}

\icmltitlerunning{Biological Evolution and Genetic Algorithms - MPhys Thesis, 2012}

\begin{document}

\twocolumn[
\icmltitle{Biological Evolution and
Genetic Algorithms \\
           Exploring the Space of Abstract Tile Self-Assembly}



\icmlsetsymbol{equal}{*}

\begin{icmlauthorlist}
\icmlauthor{Christian Schroeder de Witt}{to}
\end{icmlauthorlist}

\icmlaffiliation{to}{Department of Engineering, University of Oxford, Oxford, UK}

\icmlcorrespondingauthor{Christian Schroeder de Witt}{schroederdewitt@gmail.com}

\icmlkeywords{Machine Learning, ICML}

\vskip 0.3in
]



\printAffiliationsAndNotice{\icmlEqualContribution} 

\begin{abstract}
A physically-motivated genetic algorithm
(GA) and full enumeration for a tile-based
model of self-assembly (JaTAM) is implemented using a graphics processing unit
(GPU). We observe performance gains with
respect to state-of-the-art implementations
on CPU of factor 7.7 for the GA and 2.9
for JaTAM. The correctness of our GA
implementation is demonstrated using a
test-bed fitness function, and our JaTAM implementation is verified by classifying a
well-known search space $S_{2,8}$ based on two
tile types. The performance gains achieved
allow for the classification of a larger search
space $S^{32}_{3,8}$ based on three tile types. The
prevalence of structures based on two tile
types demonstrates that simple organisms
emerge preferrably even in complex ecosystems. The modularity of the largest structures found motivates the assumption that
to first order, $S_{2,8}$ forms the building blocks
of $S_{3,8}$. We conclude that GPUs may play
an important role in future studies of evolutionary dynamics.
\end{abstract}

\section{Introduction}
\label{introduction}

\subsection{Basic Evolutionary dynamics}

The fundamental genetic dynamics of populations of living organisms are constrained by the mechanisms of Darwinian
evolution \citep{dobzhansky_nothing_1973}. Modern science motivates the detailed investigation of evolutionary processes, e.g. designing vaccinations often requires the understanding of how pathogens will evolve in the
future \citep{domingo_quasispecies_2001}. According to Darwin,
the fitness of an individual is a measure of its fecundity, i.e. the ability of
its phenotype to reproduce \citep{darwin_c._origin_2003}. The
phenotype of an organism is the composite of all of its observable traits \citep{hartl_principles_2007}.
Neglecting environmental influences, all
information about a phenotype is encoded in the corresponding genome, or
genotype \citep{hartl_principles_2007}. Evolution may be understood as the process of adaptation of a population to peaks on a high-dimensional fitness
landscape \citep{wr32,gavrilets_evolution_1997}.

Whether adaptation occurs is governed by the ratio of exploration to exploitation exerted by the biological operators driving the traversal of
the fitness landscape: Exploration arises
due to random mutation and recombination of genomic information, exploitation
through natural selection. If exploitation
dominates, disconnected peaks may never
be discovered. If, however, exploration dominates, the trajectory through the fitness
landscape resembles a random search, unable to adapt to localised features \citep{m.a.nowak_evolutionary_2006}.
Many fitness landscapes only allow adaptation below a critical mutation rate, called
the error threshold  \citep{m.a.nowak_evolutionary_2006}.
The evolution of an individual genotype,
rather than a whole population, may be
characterised by the accessibility of its immediate fitness landscape neighbourhood
through mutation.

To this end, we define robustness to be the proportion of single mutations that don’t change phenotype, and
evolvability to be the number of different
phenotypes accessible under a single mutation \citep{wagner_robustness_2008}.
In conclusion, the key results of genetic
dynamics are encoded in the topology of
the underlying fitness landscape \citep{m.a.nowak_evolutionary_2006}.
Therefore, identifying a reasonable
genotype$\rightarrow$phenotype map is essential.
We now investigate such a map based on
the dynamics of molecular self-assembly
which is suitable for the description of the
assembling protein quaternary structures,
such as the shell of a virus \citep{johnston_exploration_2010}.

\subsection{Johnston’s Tile Assembly Model}

A ubiquitous way by which complex physical structures emerge is through the self-assembly of simple subunits \citep{whitesides_self-assembly_2002}. For
example, protein subunits assemble into
the required quaternary structures for a
particular function. Knowing how biology evolves the rule sets to self-assemble
structures is therefore crucial to understand
the evolution of simple organisms, such as viruses\citep{domingo_quasispecies_2001}. Self-assembly processes are difficult to observe experimentally due to its short length and time scales,
therefore computational simulation may be
an important source of insight  \citep{johnston_exploration_2010}.

The computational intractability and incomplete knowledge of many atomistic details motivates the development of coarsegrained models of self-assembly which attempt to capture the essential physics while
retaining tractability \citep{johnston_exploration_2010}. \citet{johnston_exploration_2010} suggest an abstract model of
self-assembly based on work by Winfree
et al \citep{rothemund_program-size_2000}. This model, which we from
now on refer to as JaTAM, exhibits rich
phenotypic phenomenology and therefore
promises to capture a lot of characteristics
of general self-assembly processes\citep{johnston_exploration_2010}.
These characteristics include the correlation
between decreasing occurence frequency
with increasing structural complexity found
in natural proteins, as well as the mutability of assembled structures \citep{ahnert_self-assembly_2009}. 

The model’s building blocks consist of squares
with labelled edges which are assembled on
a planar grid according to fixed interaction
rules. Squares are chosen from an inexhaustible library of tile types called a tile
set, of which the first tile is placed first
in the assembly process. Interaction rules
state that labels bind in pairs, i.e. $1 \leftrightarrow
2$, $3 \leftrightarrow 4$ etc. except for label $0$, which is
generally inert. Non-bonding label combinations neither repel nor attract. Binding
is thermodynamically irreversible, which is
reasonable if thermodynamic noise is much
smaller than interaction strengths between
subunits. 

Tiles may be rotated orthogonally, but not flipped. The particular choice
of interaction matrix differs from Winfree’s
original model and results in a higher ratio
of structures exhibiting dihedral, instead of
rotational, symmetry \citep{johnston_exploration_2010}. Assembly
proceeds by stochastically adding bonding
tiles to the existing structure until either no
further tiles can be assembled, or the resulting structure hits the edges of the assembly
grid.

\begin{figure}
    \centering
    \includegraphics[width=0.7\linewidth]{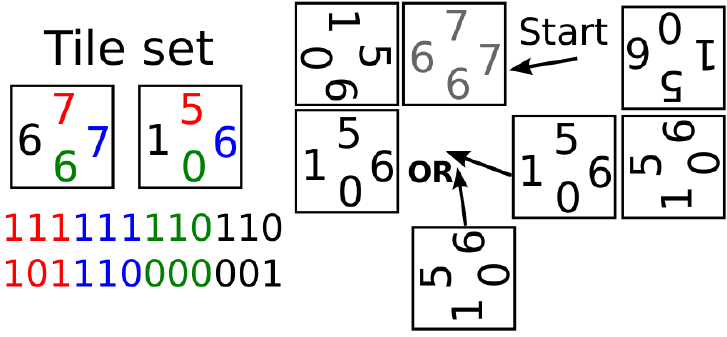}
    \caption{An example tile set that exhibits
steric non-determinism, i.e. the final structure
depends on the order in which tiles are assembled during the assembly process. The binary
string represents a simple binary encoding of the
tile set information.}
    \label{fig:tile_nondet}
\end{figure}

If a tile set is not guarantueed to always assemble to a unique grid configuration it is denoted non-deterministic. Trivial non-determinism occurs if at any step,
more than one tile can be selected to bond to
the same bonding site. More subtly, Steric
non-determinism may arise if two assembly
"arms" converge at variable speeds (see Figure \ref{fig:tile_nondet}).

Self-assembly can be cast in a formal
mathematical language, which has previously been used to e.g. prove its Turing
completeness \citep{doty_tile_2012}. For details see Appendix \ref{app:winfree}.

\subsection{The benefits of GPUs}

A graphics processing unit (GPU) is an electronic device designed for high-throughput
parallel computation at low power consumption compared to CPU clusters  \citep{nvidia_bioinformatics_2012}.
Central Processing Units (CPUs) feature
a small number of processing cores and a
large, shared low-latency cache facilitating
sequential operation. In contrast, Graphics Processing Units (GPUs) feature a large
number of processing cores with comparatively tiny low-latency caches, called shared
memory, and are designed for large dataparallel throughput.
Nvidia®, a leading GPU manufacturer,
recently announced to get involved in the
development of CPUs \citep{dally_project_2011}. This suggests
that future hardware architectures will feature intimate interaction between GPUs and
CPUs and therefore software will need to exploit the benefits of both.

The increasing need for high-throughput
computing in science  \citep{nvidia_bioinformatics_2012}, has established the use of GPUs in a number of disciplines within bioinformatics, e.g. sequence
alignment  \citep{schatz_high-throughput_2007}. In order to benefit
from GPUs, a given algorithm needs to be
efficiently parallelisable  \citep{nvidia_bioinformatics_2012}. In these
cases, the performance gains achieved compared to execution on modern CPUs are
usually on the order of one to two magnitudes  \citep{nvidia_bioinformatics_2012}.

The scope of current simulations of evolutionary dynamics performed by Johnston et
al, including JaTAM, is greatly limited by a
lack of computational performance  \citep{johnston_exploration_2010}.
The achievement of significant performance
gains for example permits the increase of
population sizes in adaptation simulations,
thus lowering error bars on simulation properties such as mean fitness. Furthermore,
it makes feasible the detailed classification
of self-assembled structures emerging from
more than two tile types, which allows the
exhaustive study of evolvability and robustness in more complex ecosystems. Both
these aspects may be required to improve
the understanding of e.g. virus evolution
\citep{domingo_quasispecies_2001}.

In this paper, we, for the first time, aim
to study a significant subset of all self-assembled JaTAM structures formed from
three tile types. To this aim, we first specify
how the tile sets assembling into phenotypes
are encoded in the respective genomes. We
then describe a class of heuristic algorithms
commonly used for the study of evolutionary dynamics of finite population sizes. Observing that these algorithms are efficiently
parallelisable, we proceed by implementing
them on GPU. Before finally classifying a
subset of all self-assembled JaTAM structures formed from three tile types, we reproduce some known results to assert the
functionality of our implementation.

\section{Implementation on GPU}

\subsection{Binary encodings}

To computationally simulate JaTAM, one
needs to somehow encode its rule sets in binary strings such that realistic biological operators may be defined. As interaction rules
are binary by nature, they can simply be
encoded in a binary matrix.
Keeping interaction rules constant in
time, all relevant phenotypic information is
encoded in the tile set. Therefore, each binary encoded tile set constitutes one genome
and the combinatorial freedom of the respective bitstring constitutes the genome search
space Sa,b, where we define a to denote the
number of tile types and b the number of
bonding labels of the underlying JaTAM assembly.

Tile sets may be encoded by sequentially
encoding the edge labels of subsequent tiles
from the tile set into a binary string, as
shown in Figure \ref{fig:tile_nondet}. If the number of different bonding labels is a power of $2$, a single
genomic mutation may be defined to consist
of a single bit flip in the respective bit string
\citep{johnston_exploration_2010}.
For a formal justification of the binary encoding chosen for this paper, see Appendix
\ref{app:enc}.
We proceed by investigating a computational algorithm which is extensively used
to simulate the evolutionary dynamics of binary encoded bitstrings \citep{holland_genetic_1992}.

\subsection{Genetic Algorithms}

Genetic Algorithms (GAs) are a class of
probabilistic search heuristics which enhance a purely random exploration by the
ability to exploit local correlations in search
space. This compromise between exploration and exploitation is efficient in locating extrema on a rugged fitness landscape \citep{holland_genetic_1992}. GAs are therefore commonly
used in problems involving complex, highdimensional fitness landscapes, from job
and process scheduling and gaming strategy
 \citep{kumar_genetic_2010} to antenna design  \citep{linden_evolutionary_2004}.

Natural selection is modelled on the basis
that the probability of a genome to be selected is proportional to the normalised fitness value of that genome. A simple selection mechanism which guarantees the above
is global roulette wheel selection with normalised weights $F_i\in\left\{\mathbb{R},F_i\geq 0\right\}$ \citep[ Appendix A]{johnston_evolutionary_2011}. $F_i=\mathcal{N}f_i$, 
, where $f_i$ is the fitness value assigned to the $i$th genome and $\mathcal{N}^{-1}=\sum\limits_{i}f_i$.
Roulette wheel selection is
equivalent to randomly throwing darts on a
dartboard sub-divided into normalised areas
of size proportional to the attached weights.
The selected individual is the one in whose
area the dart lands.
Random single-point mutation is modeled
as an independent Poisson process of mean
mutation rate $\mu$ for each gene. As far as
Johnston’s model is concerned, $\mu$ is defined
to be uniform across genomes and constant
in time. The effect of mechanism-dependent
mutation rates may be investigated by further research.

A simple way to characterise the
parametrisation of a given GA with respect
to the underlying search space are the concepts of adaptation time and discovery time
 \citep{holland_adaptation_1992}. Adaptation time measures the
mean number of generations until a proportion $x$ of the population lies within a given
fitness interval. Discovery time denotes
the mean number of generations until a
specific phenotype or fitness value has been
encountered for the first time. Both these
concepts need to be defined with respect to
well-defined initial conditions \citep{mitchell_introduction_1998}

\subsection{CUDA architecture}

Programming on General-purpose Graphics Processing Units (GPGPUs) has recently been facilitated by the introduction of novel parallel-computing frameworks, such as Nvidia’s Compute Unified
Device Architecture (CUDA)  \citep{nvidia_cuda_2012-2} and
Khronos Groups’ Open Computing Language (OpenCL) \citep{khronos_group_opencl_2012}.

The Nvidia CUDA environment(see Figure 2.3) is an abstraction layer on top
of GPU hardware architecture \citep{nvidia_cuda_2012-2}.
CUDA-code is divided into host code, which
resides on CPU caches, and device code,
which is pre-cached on GPU. The device
consists of the graphics card, the host consists of the rest of the hardware (namely the
CPU). Data transfer between host and device is low compared to device throughput.
A kernel is the GPU equivalent of a CPU
process. Device memory must be allocated
by the host prior to a kernel launch. At the
end of a simulation, results must be copied
back from GPU to CPU.
The kernel launch topology is divided
into three-dimensional grids of blocks and
threads whose dimensions are fixed for the
duration of a kernel launch.

\begin{figure}
    \centering
    \includegraphics[width=0.5\linewidth]{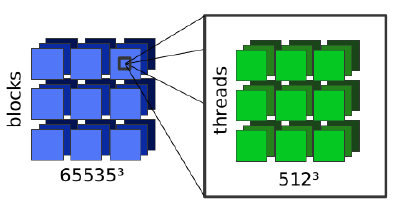}
    \caption{CUDA kernel launch topology}
    \label{fig:cuda_kernel_launch}
\end{figure}

A block may contain up to $1024$ threads,
which each execute an instance of the same
kernel. Adjacent threads within a block are
sub-divided into packages of $32$, each constituting a warp. Each thread may, to first
order, be seen to represent a single process
on an independent processing unit. There
are limits to this picture, however: threads
within a single warp are executed at lockstep, i.e. if a single thread enters a code
branch, all other threads in the same warp
have to wait until it has left the branch
again.

All threads within a block may be forced
to wait for each other at a particular code
segment by the use of an in-built barrier synchronisation function (see Figure \ref{fig:cuda_kernel_launch}). This
is useful if a code segment relies upon the
fact that all threads have terminated the
previous segment.
All threads within a block share a small
region of low-latency shared memory and a
set of fast registers. All threads of all blocks
share a large region of high-latency global
device memory (see Figure \ref{fig:cuda_kernel_launch}).


\begin{figure}
    \centering
    \begin{minipage}{1.0\linewidth}
    \includegraphics[width=\linewidth]{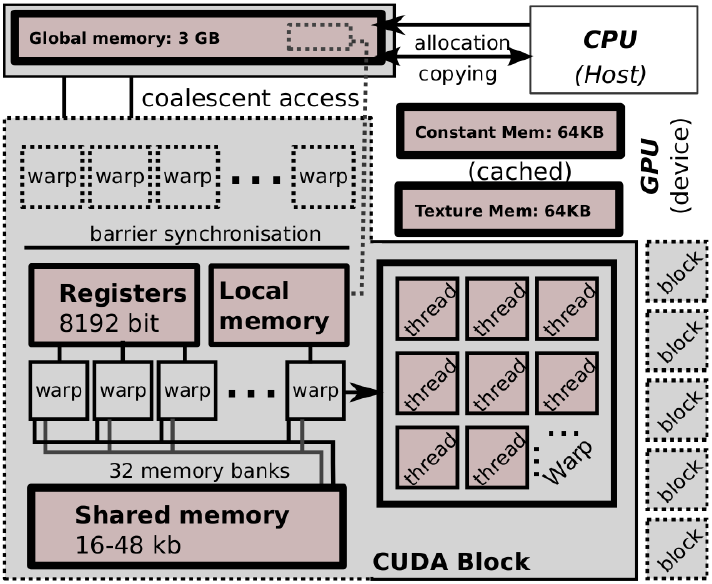}   
    \end{minipage}
    \caption{A graphical overview of NVIDIA CUDA architecture. Notice the complex memory hierarchy. }
    \label{fig:graphical_overview}
\end{figure}

\begin{figure}
    \centering
    \begin{minipage}{1.0\linewidth}
    \includegraphics[width=\linewidth]{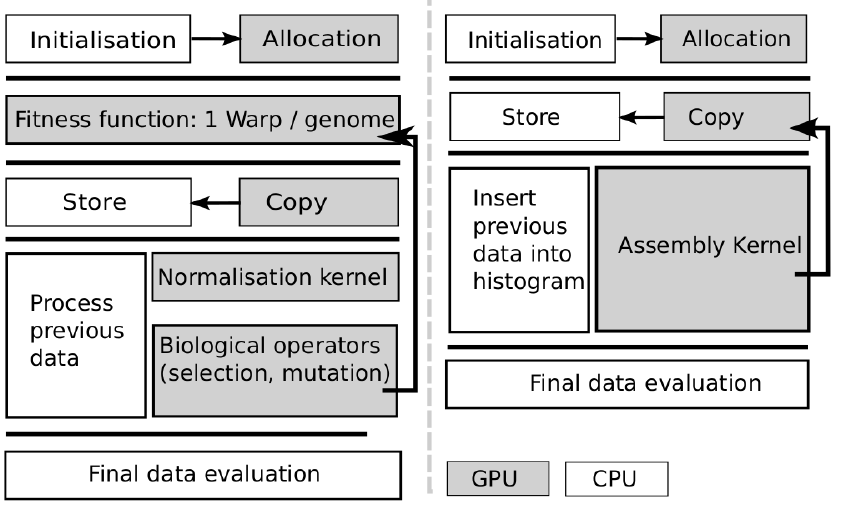} \end{minipage}
    \caption{Process layout for a model genetic algorithm with a self-assembly fitness function (on the right). Process layout for the classification of JaTAM structures (on the left).}
    \label{fig:graphical_overview}
\end{figure}

Safe simultaneous write accesses of multiple threads to shared memory resources
may be performed using atomic operations.
These are in-built routines which prevent
"race" conditions where the read-write cycles
of multiple threads are randomly mingled by
the hardware, producing an undefined outcome.
Trying to synchronize threads belonging
to different blocks via atomic operations
poses the threat of deadlocks, i.e. that
all threads wait for each other to free a
particular resource. We observe that this
form of global synchronization is most efficiently achieved by exploiting the fact that
all threads of a kernel will be synchronized
just before the kernel terminates  \citep{nvidia_cuda_2012-2}.

\section{Choice of Development Tools}

We choose to work with python as the
host language and, for performance reasons, CUDA C on device side. A templating language called Jinja2 [BSD12] serves as
a host-based metaprogramming-language,
which allows the runtime generation and
calibration of kernel code.
The CUDA C compiler, nvcc, currently
supports only a subset of the C++ specification. 
We choose to avoid recursion, function pointers and classes for
downward compatibility reasons with older
GPU architectures \citep{nvidia_cuda_2012-2}. Instead, we
use structs for object-oriented kernel design. 

For debugging purposes, we use
cuda-gdb, a Nvidia variant of the GNU
Debugger \citep{nvidia_cuda_2012-2}.
As a random number supply, we employ
the XORWOW-generator from Nvidia’s
CURAND-library \citep{nvidia_cuda_2012-1}. This generator
is similar in quality to the commonly-used
Mersenne twister  \citep{matsumoto_mersenne_1998}. CURAND may
be used with PyCUDA if name-mangling of
the C++ compiler is suppressed (see Appendix \ref{app:mangling}).
Unrolling for-loops may result in significant performance gains  \citep{murthy_optimal_2010}.

We also
observe that thread synchronization within
for-loops is sometimes unreliable. We thus
use Jinja2 to unroll loops at runtime (see
Appendix \ref{app:unroll}).

\subsection{Parallelising Johnston’s GA}

The execution of general-purpose GAs on
GPUs has shown to be especially efficient
using a "parallel island" model \citep{pospichal_parallel_2010}. For
the scope of this paper, it is sufficient to
notice that this approach by design suffers from data inconsistencies and physically
unintuitive selection operators (see Section
3.1), which is acceptable for general optimisation problems, but limits its use within
theoretical biology. At the expense of performance, We therefore adapt the model to
our needs. We assign one thread to each
genome performing the required biological
operators. In contrast to \citep{pospichal_parallel_2010}, we demand synchronisation between all threads of
all block after each generational step in order to preserve data consistency. We also introduce global roulette wheel selection both
for sexual and asexual reproduction, instead
of the local selection model proposed by
\citep{pospichal_parallel_2010}. 

To this goal, the normalisation
constant $N$ is determined using a separate
summation kernel (see Figure \ref{fig:graphical_overview}). The
roulette wheel selection is then performed
from constant device memory which is optimised for memory access patterns that
are similar for all threads within the same
warp \citep{nvidia_cuda_2012-2}.

True bit-encoding of genome information
is significantly more performant than representation by boolean values \citep{pedemonte_bitwise_2011}. Therefore, we designed efficient bit-wise implementations for mutation, single-point and
uniform crossover operators.
Importantly, mutation from Poisson distribution proves significantly faster than
conventional bit-by-bit flipping in the small
µL regime. This means we first randomly
choose the number of bits to be flipped and
subsequently choose the sites where they
should be flipped. The number of bits to
be flipped k follows a Poisson process with
mean $\lambda$ and probability distribution:
\begin{equation}
p(k)=\frac{\lambda^ ke^{-\lambda}}{k!}
\end{equation}

\begin{figure}
    \centering
    \includegraphics[width=0.8\linewidth]{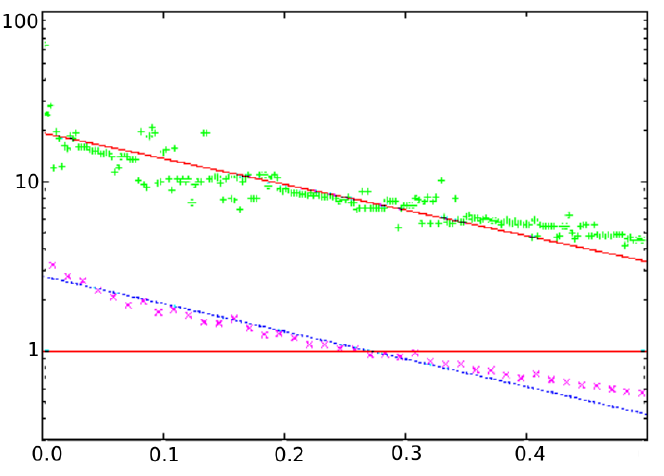}
    \caption{Performance ratio for bitwise mutation / mutation by distribution vs. mutation rate $\mu L$.}
    \label{fig:bitwise_mut}
    \vspace{-2.5em}
\end{figure}

A bitmask is used to avoid multiple flipping
at the same sites but may, for $\mu L \ll 1$, be
omitted. Figure \ref{fig:bitwise_mut} below demonstrates a
significant speed-up even at small genome
bytelengths for $\mu L \leq 0.5$. The graph on top
corresponds to a genome length of $128$ bytes,
the lower graph to a length of $3$ bytes. $105$
iterations using a single CPU thread were
performed for each data point. For large
genome bytelengths, mutation from distribution is several orders of magnitude faster
than bit-wise flipping.
For our efficient bit-wise implementations
of common evolutionary operators see appendices D, D.1.

\subsection{Parallelising JaTAM}

We choose to assemble $32$ tile sets per block,
scheduling $8$ threads from adjacent warps
per assembly. This allows us to store the
grid for each assembly within a single shared
memory bank, enabling an efficient shared
memory access pattern called bank-conflict
free access \citep{nvidia_cuda_2012}. Ensuring that assembly threads are from different warps allows
the usage of atomic operations within each
assembly process - atomic operations among
threads of the same warps result in undefined behaviour as these threads execute at
lockstep  \citep{nvidia_cuda_2012-2}.
Grid initialisation is performed by all the
threads in an assembly process in parallel.
This compensates for shared memory access
latency.

For the polyomino-assembly fitness function, we a single-threaded movelist-based algorithm offered by Johnston, which stochastically grazes through the assembly frontier
on a finite grid of dimension $d$  \citep{johnston_exploration_2010}. This
implies that $7$ of the $8$ assembly threads
are idle during assembly. Each position on
the assembly frontier is isotropically checked
for ambiguous bonding situations and in
this case, the structure is classified nondeterministic. This checks for trivial and
first-order steric non-determinism at the
same time and we therefore do not distinguish between the two. Higher-order steric
non-determinancy is stochastically detected
by detailed comparison of k independent assembly processes. See Appendix \ref{app:movelist} for a
flow chart of the movelist assembly process.

To optimise the assembly process, each
bonding label is pre-assigned a list of tiles
and their orientations which bond to it.
In order to only detect phenotypic nondeterminancy, we remark that this list needs
to be further processed in order to eliminate
double tile types and symmetric tiles. Also,
steric non-determinism detection should be
conducted only after the assembly has terminated to allow rotational invariance of
assembled shapes - this could be included
straightforwardly, but we prefer to be consistent with \citet{johnston_evolutionary_2011}.

As shared memory features no heap for
dynamic memory allocation, we choose to
implement the movelist as a combination of
a pre-allocated \textit{Last-In-First-Out (LIFO)}-
stack and grid markings (see below).
This implies a higher degree of locality in
assembly frontier traversal compared to
random traversal using dynamic lists and
reduces the need for pre-allocated storage
space by about $5$ 
We, however, observe that the detection of
steric non-determinism is not significantly
affected by localised traversal as movelist
elements are added to the stack in random
oder.

\begin{figure}
    \centering
    \includegraphics[width=0.8\linewidth]{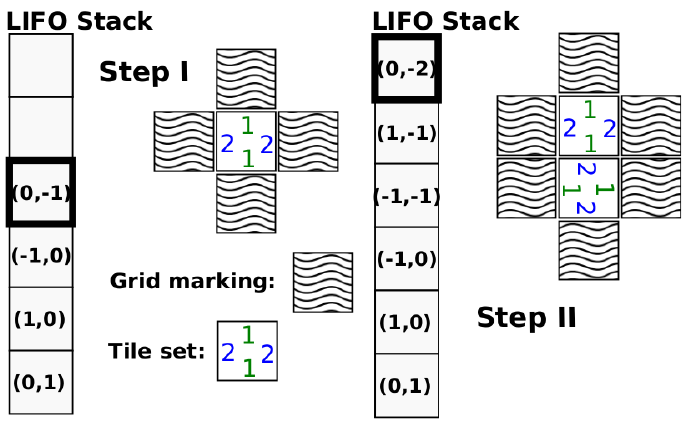}
    \caption{Illustration of an assembly process using
both grid markings and a LIFO-stack.
Note that elements are ramdomly inserted
and removed from the top, therefore,
growth exhibits a large degree of spatial
locality.}
    \label{fig:lifo}
\end{figure}

\subsection{Classifying self-assembled shapes}

Grid-wise comparisons of assembly structures as employed by Johnston are unsuitable for the classification of larger search
spaces than $S_{2,8}$ as they are resourceexpensive and ambiguous in higher search
spaces \citep{johnston_exploration_2010}. As the register manipulation of composite data types in CUDA
is tedious, I employ a $32$-bit \textit{one-at-a-time hash} \citep{jenkins_hash_1997}. Each phenotype shape is
hashed by subsequently encoding the positions of all occupied grid sites relative to the
coordinates of an assembly grid cropped to
the phenotype shape. Cropping is readily
parallelisable using atomic operations (see
Appendix \ref{app:cropping}). For an expected number
of $1000$ different phenotypes, the probability
that at least one hash collision occurs is

\begin{equation}
    1 - \prod\limits_{i=0}^{1000} \left(1-i\cdot 2^{-32}\right) \approx 0.01\%.
\end{equation}

Therefore, a $32$-bit hash is sufficient for
the purpose of this paper. We note that
hashes may be made rotationally invariant
if one chooses to hash each grid for all
four orthogonal rotations individually,
sorts these hashes by e.g. numerically
and sequentially hashes them again with
Jenkin’s hash function.

\begin{figure}
    \centering
    \includegraphics[width=0.8\linewidth]{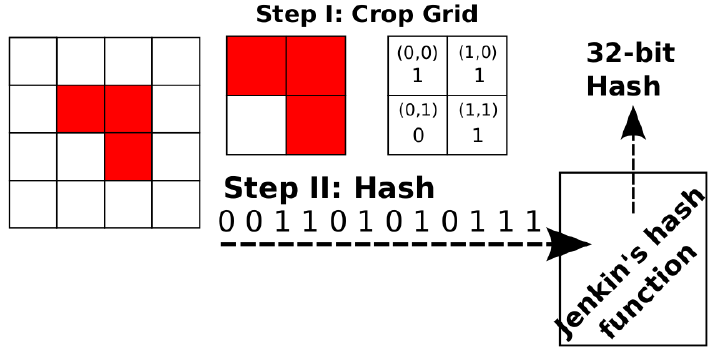}
    \caption{An illustration of grid hashing.}
    \label{fig:grid_hashing}
\end{figure}

\subsection{Exploiting CPU and GPU concurrently}

Classifying a genome search space consists
of assembling each tile set, hashing the resulting structure and inserting it into a histogram counting the occurrences of each
hash. As GPUs have a finite global memory size, on a Tesla $C2050$, only about $2^16$
assembly grids of dimension $19\times 19$ may
be stored at any one time. For searchspace classification, this requires execution
in batches. 

However, the execution may
be asynchronous: While the GPU executes
one batch, the previous batch is meanwhile
sorted on CPU (see Figure \ref{fig:graphical_overview}). For our
model, this was shown to efficiently hide
CPU processing times. Alternatively, one
may just copy the hashes from GPU and
the bitstrings for the respective tile sets assembling to those hashes. After having created the histogram on CPU, one may just
re-assign assembled structures to hashes by
considering the structure which predominantly assembles from the respective tile set.

\section{Results}

All CPU timing references were measured
by Johnston \citep{johnston_exploration_2010} on an Intel\textsuperscript{\tiny\textregistered} Core\textsuperscript{TM} i$5$-
$2520M$ CPU running at $2.50GHz$ and sufficient RAM. GPU timings were measured
on a single NVidia Tesla $C2050$ card. Both
these devices represent state-of-the-art technology.

\subsection{Classification of $S_{2,8}$}

To verify the correctness of the polyomino
assembly routine, I re-classify the search
space $S_{2,8}$. Comparing results to Johnston’s
yields identical values for the occurrence
of deterministic structures (see Figure \ref{fig:s28_enumeration}).
There is evidence that the mis-classification
of non-deterministic structures as deterministic structures can be made arbitrarily small with increasing k (see Figure \ref{fig:s28_enumeration}
). The only phenotype shape with both deterministic and sterically non-deterministic
rulesets is found to be the $12$-mer (see Figure
\ref{fig:s28_enumeration}). 

Structures of steric non-determinism
outweigh unbound structures by a factor of
$\approx2.2$. This suggests to optimise our assembly algorithm for steric non-determinism detection in the future, which may be achieved
by conducting as many assembly processes
in parallel at any one time. Finally, a speedup of a factor $2.9$, as compared to Johnston, was detected for our CPU/GPU hybrid implementation.

\subsection{Genetic algorithms on a Fujiyama landscape}

To demonstrate the flawnessless of our implementation, we implement our genetic algorithm on a test-bed fitness landscape referred to as `Fujiyama'-landscape. This
landscape consists of a single, broad peak.
In evolutionary dynamics, the study of
adaptation behaviour on Fujiyama landscapes is insightful as locally, any smooth
fitness function may be approximated by a
single peak. In simple binary encoding, a
suitable fitness function consists of the Hamming weight $H(x)$ of a genome $x$, which is
equivalent to the number of set bits in the
respective bitstring. We choose to investigate the adaptation and discovery times for
a population size of $512$. 

Each data point in Figure \ref{fig:adaptation_times} represents the median of $100$ independent simulations. A simulational cutoff at $20000$ generations is employed. Error
bars were derived from bootstrapping with
sample size $100$ and $104$
repetitions.
I find discovery times to power-law decrease with increasing mutation rate, and
the existence of a catastrophic adaptation
cut-off in the vicinity of $\mu L = 1$. Both
these results are fundamental characteristics
of genetic algorithms \citep{johnston_exploration_2010}. A speed-up
of our implementation of factor $7.7$, as compared to Johnston, was measured.

\subsection{Classification of a subset $S^{32}_{3,8}$ of $S_{3,8}$}

\begin{figure*}
    \centering
    \includegraphics[width=1.0\linewidth]{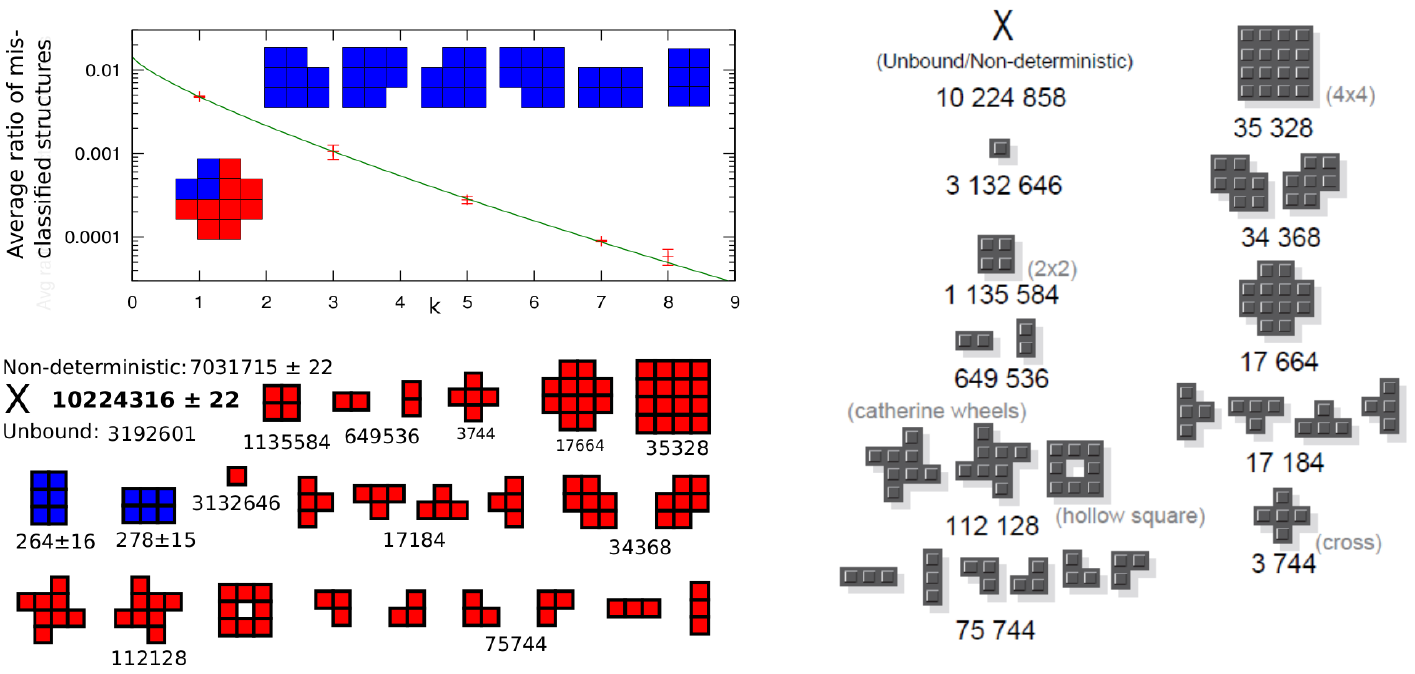}
    \caption{Right: Johnston’s enumeration of $S_{2,8}$. Upper left: Ratio of mis-classified structures with $k$. Lower left: GPU classification of $S_{2,8}$ at $k=8$.}
    \label{fig:s28_enumeration}
\end{figure*}

\begin{figure}
    \centering
    \includegraphics[width=0.9\linewidth]{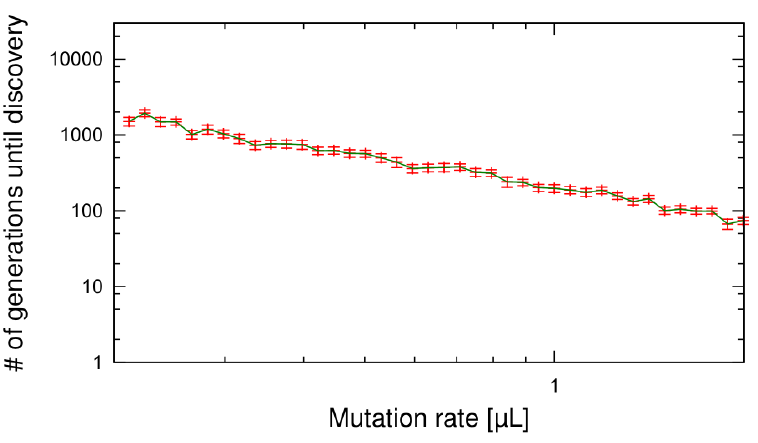}
    \caption{Discovery times for $H=25$ on Fujiyama landscape \citep{johnston_exploration_2010}.}
    \label{fig:discovery_times}
\end{figure}

\begin{figure}
    \centering
    \includegraphics[width=0.9\linewidth]{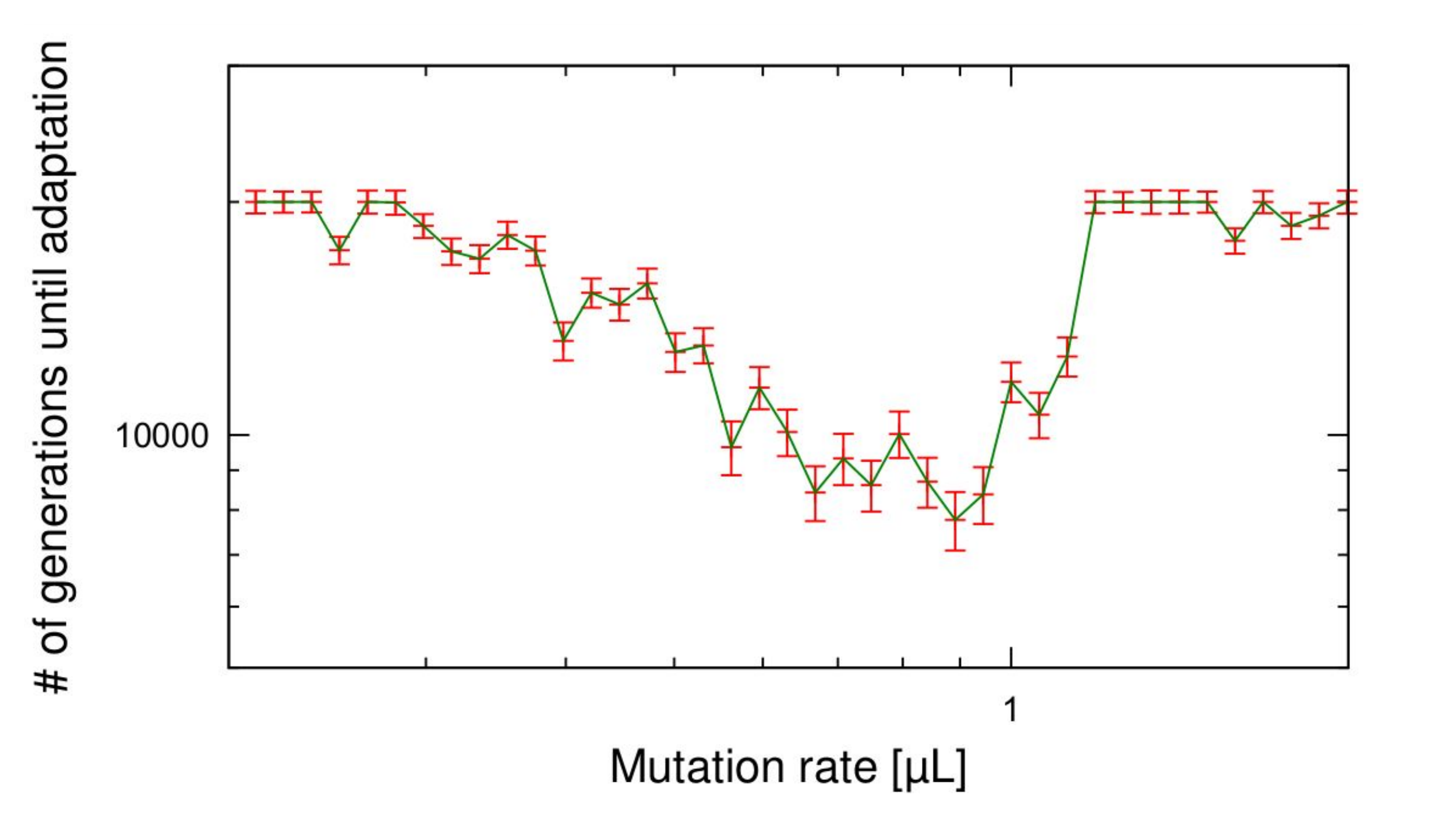}
    \caption{Adaptation times for half of the population to be at $H\geq 25$ on Fujiyama landscape}
    \label{fig:adaptation_times}
\end{figure}

Whereas there exists a large amount of literature on the study of individiual search
spaces  \citep{johnston_exploration_2010}, this is currently not the case
for the comparison between different search
spaces. Using our CPU/GPU hybrid implementation of JaTAM and a process layout given in Figure \ref{fig:graphical_overview}, we therefore for the
first time classify a search space which may
be expected to yield phenomenology distinct
from $S_{2,8}$. The JaTAM search space for $3$
tile types and $8$ distinct bonding labels is
4096-times larger than $S_{2,8}$ and its classification is expected to yield a rich amount of
shape phenomenology. 

We provide a GPU
classification of the first $32$ bits (thus a $16$th)
of this search space, which means that the
seed tile type and second tile type may have
arbitrary bonding labels, while the third tile
type is fixed to bonding labels $\{x_1,x_2,y,0\}$,
where $x_1, x_2 \in [0,7]$ and $y\in\{2,4,6\}$. A frequency classification of deterministic structures shows that the deterministic structures of $S_{2,8}$ are the most frequent ones in
$S^{32}_{3,8}$ \ref{fig:s28_enumeration}. 

This is expected, as the probability that the third tile does not interact with the other two tiles is increased due to the
fixed zero bonding site. The frequency differences of certain flipped forms is caused by
the fact that these would require tile three
to flip bonding labels, which is only possible
if $x_2=0$ or $y=x_1$.

The most frequently encountered structures in $S_{3,8}$ are in fact the simple structures found in $S_{2,8}$. This indicates that
simple structures continue to be common
in complex ecosystems. Furthermore, the
largest structures formed exhibit striking
modularity: Most of them consist out of
trivially-assembled copies of members of
$S_{2,8}$. This motivates the assumption that
to first order, $S_{2,8}$ forms the building blocks
of $S_{3,8}$. An overview of all $361$ deterministic
structures encountered in $S_{3,8}$ can be found
in Appendix A.

\begin{figure*}
    \centering
    \includegraphics[width=1.0\linewidth]{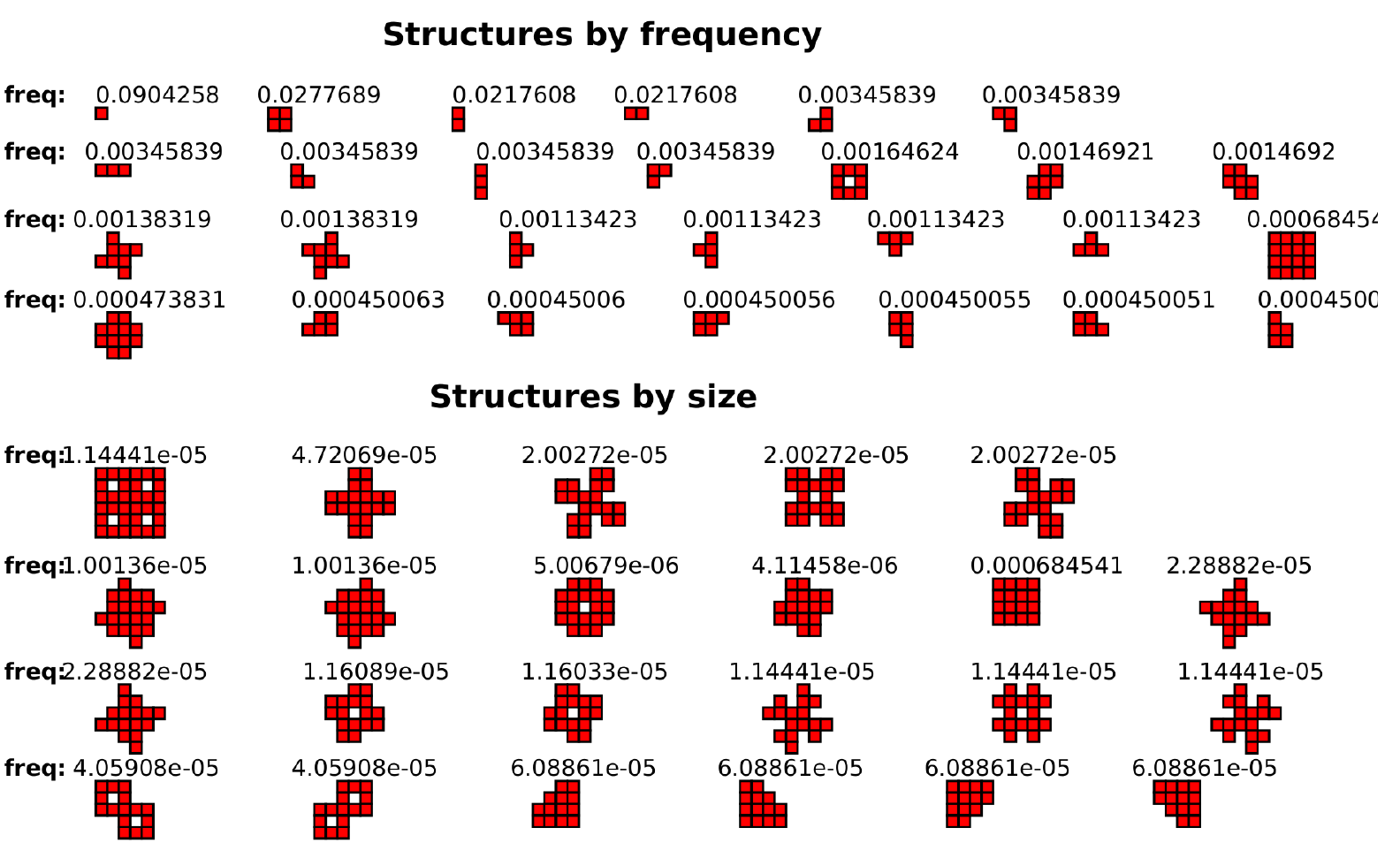}
    \caption{Selected phenotype shapes from the classification of $S^{32}_{3,8}$ at $k=7$. For a full list, see Appendix \ref{app:s38}}
    \label{fig:selected_s38}
\end{figure*}

\section{GPU: Pitfalls, Performance Gains and Coding Standards}

\subsection{Pitfalls}

Langdon advised in 2011 that scientists
should refrain from writing new pieces of
software using CUDA whenever possible because of the frequent appearance of bugs
that were hard or almost impossible to
debug  \citep{langdon_debugging_2011}. As this project featured
rather complex kernels (up to 1706 lines for
JaTAM classification, which is at least 17-
times larger than typical kernel sizes according to Langdon \citep{langdon_debugging_2011}), virtually all of the
pitfalls reported by Langdon were encountered. Some of these resulted in intraceable
bugs and lead three times to a complete redesign of our source code methodology.

On top of Langdon’s observations, compiler errors generated by the CUDA compiler nvcc were often highly misleading (if it
hadn’t crashed during compilation). Tracebacks of cuda-gdb often merely featured unresolved memory addresses and printf functions were prone to buffer overflows, thus
making debugging frequently unfeasible and
turning the graphics card into an effective
"black box".

In response, we developed a coding model
and compatibility layer that allowed us to
first develop and debug our multi-threaded
kernel code on CPU. Then, we performed
minor modifications to port the code to
GPU architecture. This approach in principle also allows easy performance comparisons to implementations on multi-threaded
CPUs.

\subsection{Opportunities for further code optimisation}

The complexities of code development made
it impossible to include some important optimisation features in the current version
of our implementation. This may explain
why the performance gains achieved for the
GA are not as large as may have been expected from the literature  \citep{pospichal_parallel_2010}. Most
notably, our current JaTAM classification
spends 83\% of its time on assembly grid
initialisation. We suggest that this time
may be reduced by keeping track of the
positions were tiles have been assembled
and only re-initialising these. A

Additionally, we found that grid initialisation times
could be reduced up to $4$-fold by increasing
memory access granularity through the use
of pointer re-interpretation (see Appendix
\ref{app:reinterpret}). Time spent assembling the structures themselves may be reduced using a
parallel assembly process, which may either
involve a pure random pick in-place assembly routine (see the flowchart in Appendix
\ref{app:flowchart}) or a hybrid approach featuring either
separate or shared movelists for each assembly thread (see Appendix \ref{app:lifo} for a push function of a multi-threading safe LIFO stack).

We conclude that even our only modestly optimised GPU implementations yield
significant performance gains compared
to state-of-the-art CPU implementations.
These performance gains may be further increased by performing further optimisation
measures as suggested. This indicates that
GPUs are likely to have a prominent future
in the study of genetic population dynamics.

\subsection{Coding standards in the sciences}

The difficulties encountered in the development of our GPU implementation immediately raise the question whether it is possible
to avoid many common pitfalls by employing state-of-the-art coding standards from
within the computer science community. We
come up with the following suggestions:
Software written in imperative programming languages should be designed from
flow-charts (procedural) and/or UML activity diagrams [see App. \ref{app:uml}] (object-oriented)
level. These charts should be supplied for
peer-review and enable flow-level analysis.

The source code of unit tests (see Appendix
\ref{app:uml}) and their results should be supplied for
peer-review as well to complement flow-level
analysis.
Domain-specific languages (DSLs) are
special-purpose programming languages designed for usage in particular problem domains. Usage of these may abstract away
many possible error sources. Creating a new
DSL, however, is comparable to writing a
compiler and thus difficult and error-prone
 \citep{cartney_generating_2012}. An example of a system that allows
the creation of new DSLs is Cartneys recent
work  \citep{cartney_generating_2012}.
Functional programming languages allow
for an abstract flow-independent problem
formulation, often resulting in smaller code
sizes as well. There exists a large infrastructure of quantitative unit testing tools such
as QuickCheck. Equational reasoning allows
for fine-grained formal verification in some
circumstances  \citep{harper_interview_2012}.
Problems that are efficiently parallelisable
should be expressed in a concurrent programming language. 

While a number of
concurrent imperative languages are under
development, such as python/copperhead
 \citep{copperhead_copperhead_2012}, functional programming languages
arguably embody parallelism more naturally  \citep{chakravarty_general_2009}. The functional programming language Haskell offers industry-grade
extensions for data-parallel programming
 \citep{wu_oxford_2012,ghc_data-parallel_2012}.
Computer scientists are currently developing functional programming language
derivates, such as Accelerate  \citep{chakravarty_accelerating_2011},
which are compiled to CUDA C or Parallel Thread eXecution  \citep{nvidia_ptx:_2007} code and thus
produce code for later execution on GPUs.

We conclude that future researchers in the
sciences may especially benefit from learning concurrent, as well as functional, programming languages.
For performance reasons, we, for this
project, chose to program in CUDA C and
Python, which are both imperative languages. 

\section{Conclusion and Outlook}

We implemented a biologically-inspired evolutionary model on a graphics card. This
resulted in a performance gain of factor
7.7 over state-of-the-art implementations.
The correctness of our implementation was
demonstrated on a Fujiyama fitness landscape. We developed a novel mutation algorithm in the process which significantly
outperformed conventional algorithms at
low mutation rates. 

Additionally, a two-dimensional model of self-assembly was implemented using a CPU/GPU hybrid model.
This resulted in a performance gain of factor $2.9$ over state-of-the-art implementations. The correctness of our implementation was demonstrated by classifying a well-known search space named $S_{2,8}$. A universal
method of classifying self-assembled shapes
using hashing was introduced. We list additional optimisation measures which may result in further significant performance gains
in the future.

Using our CPU/GPU hybrid implementation, a subset of a previously unstudied self-assembly search space named $S_{3,8}$ was classified. It is found that this search space prevalently contains simple structures, which indicates that simple organisms preferably
emerge even in complex ecosystems. The
modularity of the largest structures found
motivates the assumption that to first order, $S_{2,8}$ forms the building blocks of $S_{3,8}$.
Our work may easily be extended to study
evolvability and robustness in $S_{3,8}$.

Finally, on the basis of this project,
we discuss the challenges that novel hardware architecture poses to traditional coding
standards in the sciences and what technologies may be taught to future students in the
sciences in order to increase the tractability
of their code.

We conclude that GPUs may have significant applications in future studies of evolutionary dynamics in theoretical condensed
matter physics.

\Urlmuskip=0mu plus 1mu\relax
\bibliography{main}

\begin{thebibliography}{46}
\providecommand{\natexlab}[1]{#1}
\providecommand{\url}[1]{\texttt{#1}}
\expandafter\ifx\csname urlstyle\endcsname\relax
  \providecommand{\doi}[1]{doi: #1}\else
  \providecommand{\doi}{doi: \begingroup \urlstyle{rm}\Url}\fi

\bibitem[Ahnert et~al.(2009)Ahnert, Johnston, Fink, Doye, and
  Louis]{ahnert_self-assembly_2009}
Ahnert, S.~E., Johnston, I., Fink, M., Doye, J., and Louis, A.
\newblock Self-assembly, modularity and physical complexity.
\newblock \emph{Phys. Rev. E}, 2009.

\bibitem[Cartney(2012)]{cartney_generating_2012}
Cartney, L.
\newblock Generating {GPU} code from domain specific languages, February 2012.

\bibitem[Chakravarty(2009)]{chakravarty_general_2009}
Chakravarty, M.
\newblock General purpose {GPU} computing @ programming languages \& systems,
  2009.
\newblock URL
  \url{http://www.cse.unsw.edu.au/~pls/cuda-workshop09/slides/gpu-research.pdf}.

\bibitem[Chakravarty et~al.(2011)Chakravarty, Keller, Lee, {McDonell}, and
  Grover]{chakravarty_accelerating_2011}
Chakravarty, M., Keller, G., Lee, S., {McDonell}, T., and Grover, V.
\newblock Accelerating haskell array codes with multicore {GPUs}.
\newblock \emph{Declarative Aspects of Multicore Programming {(DAMP} 2011)},
  2011.

\bibitem[copperhead(2012)]{copperhead_copperhead_2012}
copperhead.
\newblock copperhead - data parallel python.
\newblock http://code.google.com/p/copperhead/, 2012.
\newblock URL \url{http://code.google.com/p/copperhead/}.

\bibitem[Dally(2011)]{dally_project_2011}
Dally, B.
\newblock {"Project} denver" processor to usher in new era of computing.
\newblock
  http://blogs.nvidia.com/2011/01/project-denver-processor-to-usher-in-new-era-of-computing/,
  January 2011.
\newblock URL
  \url{http://blogs.nvidia.com/2011/01/project-denver-processor-to-usher-in-new-era-of-computing/}.

\bibitem[Darwin(2003)]{darwin_c._origin_2003}
Darwin, C.
\newblock \emph{Origin of Species}.
\newblock Signet Classic, 2003.

\bibitem[Dobzhansky(1973)]{dobzhansky_nothing_1973}
Dobzhansky, T.
\newblock Nothing in biology makes sense except in the light of evolution.
\newblock \emph{The American Biology Teacher}, 35:\penalty0 125--129, March
  1973.

\bibitem[Domingo et~al.(2001)Domingo, Holland, and
  Biebricher]{domingo_quasispecies_2001}
Domingo, E., Holland, J.~J., and Biebricher, C.~K.
\newblock \emph{Quasispecies and {Rna} {Virus} {Evolution}: {Principles} and
  {Consequences}}.
\newblock R G Landes Co, Georgetown, Tex. : Austin, Tex, March 2001.
\newblock ISBN 978-1-58706-010-6.

\bibitem[Doty et~al.(2012)Doty, Lutz, Patitz, Schweller, Summers, and
  Woods]{doty_tile_2012}
Doty, D., Lutz, J.~H., Patitz, M.~J., Schweller, R.~T., Summers, S.~M., and
  Woods, D.
\newblock The {Tile} {Assembly} {Model} is {Intrinsically} {Universal}.
\newblock pp.\  302--310. IEEE Computer Society, October 2012.
\newblock ISBN 978-1-4673-4383-1.
\newblock \doi{10.1109/FOCS.2012.76}.
\newblock URL
  \url{https://www.computer.org/csdl/proceedings-article/focs/2012/4874a302/12OmNyuPKTx}.
\newblock ISSN: 0272-5428.

\bibitem[Gavrilets(1997)]{gavrilets_evolution_1997}
Gavrilets, S.
\newblock Evolution and speciation on holey adaptive landscapes.
\newblock \emph{Trends in Ecology \& Evolution}, 12\penalty0 (8):\penalty0
  307--312, August 1997.
\newblock ISSN 0169-5347.
\newblock \doi{10.1016/S0169-5347(97)01098-7}.
\newblock URL
  \url{https://www.sciencedirect.com/science/article/pii/S0169534797010987}.

\bibitem[{GHC}(2012)]{ghc_data-parallel_2012}
{GHC}.
\newblock {Data-Parallel} haskell, 2012.
\newblock URL
  \url{http://www.haskell.org/haskellwiki/GHC/Data_Parallel_Haskell}.

\bibitem[Group(2012)]{khronos_group_opencl_2012}
Group, K.
\newblock {OpenCL} - the open standard for parallel programming of
  heterogeneous systems, March 2012.
\newblock URL \url{http:/www.khronos.org/opencl/}.

\bibitem[Harper(2012)]{harper_interview_2012}
Harper, T.
\newblock Interview on software verification, January 2012.

\bibitem[Hartl \& Clark(2007)Hartl and Clark]{hartl_principles_2007}
Hartl, D. and Clark, A.
\newblock \emph{Principles of Population Genetics}.
\newblock Sinauer Associates, Inc. Publishers, Sunderland, {MA}, 4th edition,
  2007.
\newblock ISBN 978-0-878-93308-2.

\bibitem[Holland(1992{\natexlab{a}})]{holland_adaptation_1992}
Holland, J.
\newblock \emph{Adaptation in natural and artificial systems}.
\newblock {MIT} press Cambridge, {MA}, 1992{\natexlab{a}}.

\bibitem[Holland(1992{\natexlab{b}})]{holland_genetic_1992}
Holland, J.~H.
\newblock Genetic {Algorithms}.
\newblock \emph{Scientific American}, 267\penalty0 (1):\penalty0 66--73,
  1992{\natexlab{b}}.
\newblock ISSN 0036-8733.
\newblock URL \url{https://www.jstor.org/stable/24939139}.
\newblock Publisher: Scientific American, a division of Nature America, Inc.

\bibitem[Jenkins(1997)]{jenkins_hash_1997}
Jenkins, B.
\newblock Hash functions.
\newblock 1997.

\bibitem[Johnston(2010)]{johnston_exploration_2010}
Johnston, I.
\newblock \emph{Exploration, Exploitation \& Complexity in Biological Evolution
  and {Self-Assembly}}.
\newblock doctoral thesis, University of Oxford, Oxford, {UK}, 2010.

\bibitem[Johnston et~al.(2011)Johnston, Ahnert, Louis, and
  Doye]{johnston_evolutionary_2011}
Johnston, I., Ahnert, S.~E., Louis, A., and Doye, J.
\newblock Evolutionary dynamics in a simple model of {Self-Assembly}.
\newblock \emph{Phys. Rev. E}, 2011.
\newblock \doi{10.1103/PhysRevE.83.066105}.

\bibitem[Kumar et~al.(2010)Kumar, Husian, Upreti, and
  Gupta]{kumar_genetic_2010}
Kumar, M., Husian, M., Upreti, N., and Gupta, D.
\newblock {GENETIC} {ALGORITHM:} {REVIEW} {AND} {APPLICATION}.
\newblock \emph{International Journal of Information Technology and Knowledge
  Management}, 2\penalty0 (No. 2):\penalty0 451--454, December 2010.

\bibitem[Langdon(2011)]{langdon_debugging_2011}
Langdon, W.
\newblock Debugging {CUDA}.
\newblock \emph{{GECCO’11}}, 2011.
\newblock
  \doi{http://delivery.acm.org/10.1145/2010000/2002028/p415-langdon.pdf}.

\bibitem[Linden et~al.(2004)Linden, Lohn, Hornby, and
  Kraus]{linden_evolutionary_2004}
Linden, D., Lohn, J., Hornby, G., and Kraus, W.
\newblock Evolutionary design of an x-band antenna for {NASA's} space
  technology 5 mission.
\newblock \emph{Antennas and Propagation Society International Symposium, 2004.
  {IEEE}}, 3:\penalty0 2313 -- 2316, 2004.

\bibitem[{M.A.Nowak}(2006)]{m.a.nowak_evolutionary_2006}
{M.A.Nowak}.
\newblock \emph{Evolutionary Dynamics}.
\newblock 2006.

\bibitem[Matsumoto \& Nishimura(1998)Matsumoto and
  Nishimura]{matsumoto_mersenne_1998}
Matsumoto, M. and Nishimura, T.
\newblock Mersenne twister: A 623-dimensionally equidistributed uniform
  pseudorandom number generator.
\newblock \emph{{ACM} Transactions {onModeling} and Computer Simulation},
  8(1):\penalty0 3–30, January 1998.

\bibitem[Mitchell(1998)]{mitchell_introduction_1998}
Mitchell, M.
\newblock \emph{A Introduction to Genetic Algorithms}.
\newblock The {MIT} Press, 1998.

\bibitem[Murthy et~al.(2010)Murthy, Ravishankar, Baskaran, and
  Sadayappan]{murthy_optimal_2010}
Murthy, G.~S., Ravishankar, M., Baskaran, M.~M., and Sadayappan, P.
\newblock Optimal loop unrolling for {GPGPU} programs.
\newblock In \emph{2010 {IEEE} {International} {Symposium} on {Parallel}
  {Distributed} {Processing} ({IPDPS})}, pp.\  1--11, April 2010.
\newblock \doi{10.1109/IPDPS.2010.5470423}.
\newblock ISSN: 1530-2075.

\bibitem[{NVidia}(2007)]{nvidia_ptx:_2007}
{NVidia}.
\newblock {PTX:} parallel thread execution, October 2007.
\newblock {ISA} Version 1.1.

\bibitem[{NVidia}(2012{\natexlab{a}})]{nvidia_bioinformatics_2012}
{NVidia}.
\newblock Bioinformatics and life sciences {(GPU)}, 2012{\natexlab{a}}.
\newblock URL \url{http://www.nvidia.com/object/bio_info_life_sciences.html}.

\bibitem[{NVidia}(2012{\natexlab{b}})]{nvidia_cuda_2012}
{NVidia}.
\newblock {CUDA} c best practices guide, January 2012{\natexlab{b}}.
\newblock URL
  \url{http://developer.download.nvidia.com/compute/DevZone/docs/html/C/doc/CUDA_C_Best_Practices_Guide.pdf}.

\bibitem[{NVidia}(2012{\natexlab{c}})]{nvidia_cuda_2012-1}
{NVidia}.
\newblock {CUDA} toolkit 4.1: {CURAND} guide, January 2012{\natexlab{c}}.
\newblock URL
  \url{http://developer.download.nvidia.com/compute/DevZone/docs/html/CUDALibraries/doc/CURAND_Library.pdf}.

\bibitem[{NVidia}(2012{\natexlab{d}})]{nvidia_cuda_2012-2}
{NVidia}.
\newblock {CUDA} c programming guide, 2012{\natexlab{d}}.
\newblock URL
  \url{http://developer.download.nvidia.com/compute/DevZone/docs/html/C/doc/CUDA_C_Programming_Guide.pdf}.

\bibitem[Pedemonte et~al.(2011)Pedemonte, Alba, and
  Luna]{pedemonte_bitwise_2011}
Pedemonte, M., Alba, E., and Luna, F.
\newblock Bitwise operations for {GPU} implementation of genetic algorithms.
\newblock \emph{Proceedings of the 13th annual conference companion on Genetic
  and evolutionary computation}, 2011.
\newblock \doi{10.1145/2001858.2002031}.

\bibitem[Pospichal et~al.(2010)Pospichal, Jaros, and
  Schwarz]{pospichal_parallel_2010}
Pospichal, P., Jaros, J., and Schwarz, J.
\newblock Parallel genetic algorithm on the {CUDA} architecture.
\newblock Brno, 2010. Masaryk University.
\newblock ISBN 978-80-87342-10-7.
\newblock URL \url{http://www.fit.vutbr.cz/research/view_pub.php?id=9432}.

\bibitem[Ronacher(2012)]{ronacher_jinja2_2012}
Ronacher, A.
\newblock Jinja2, 2012.
\newblock URL \url{http://jinja.pocoo.org/docs/}.

\bibitem[Rothemund \& Winfree(2000)Rothemund and
  Winfree]{rothemund_program-size_2000}
Rothemund, P. and Winfree, E.
\newblock The program-size of self-assembled squares (extended abstract).
\newblock 2000.

\bibitem[Schatz et~al.(2007)Schatz, Trapnell, Delcher, and
  Varshney]{schatz_high-throughput_2007}
Schatz, M., Trapnell, C., Delcher, A., and Varshney, A.
\newblock High-throughput sequence alignment using graphics processing units.
\newblock \emph{{BMC} Bioinformatics}, \penalty0 (8):\penalty0 474, 2007.
\newblock \doi{10.1186/1471-2105-8-474}.

\bibitem[Schmielau(2012)]{schmielau_implementing_2012}
Schmielau, T.~t.
\newblock Implementing a {FIFO}-list in shared memory using atomics, 2012.
\newblock URL
  \url{https://forums.developer.nvidia.com/t/implementing-a-fifo-list-in-shared-memory-using-atomics-memory-inconsistencies-each-thread-in-a-wa/25864}.

\bibitem[Shannon(1948)]{shannon_mathematical_1948}
Shannon, C.
\newblock A mathematical theory of communication.
\newblock \emph{Bell System Technical Journal}, 27:\penalty0 379–423, 1948.

\bibitem[Soloveichik \& Winfree(2007)Soloveichik and
  Winfree]{soloveichik_complexity_2007}
Soloveichik, D. and Winfree, E.
\newblock Complexity of {Self-Assembled} shapes.
\newblock \emph{{SIAM} J. Comput.}, \penalty0 (36):\penalty0 1544--1569, 2007.

\bibitem[Wagner(2008)]{wagner_robustness_2008}
Wagner, A.
\newblock Robustness and evolvability: a paradox resolved.
\newblock \emph{Proceedings. Biological Sciences}, 275\penalty0
  (1630):\penalty0 91--100, January 2008.
\newblock ISSN 0962-8452.
\newblock \doi{10.1098/rspb.2007.1137}.

\bibitem[Wang(1961)]{wang_proving_1961}
Wang, H.
\newblock Proving theorems by pattern recognition — {II}.
\newblock \emph{The Bell System Technical Journal}, 40\penalty0 (1):\penalty0
  1--41, January 1961.
\newblock ISSN 0005-8580.
\newblock \doi{10.1002/j.1538-7305.1961.tb03975.x}.
\newblock Conference Name: The Bell System Technical Journal.

\bibitem[Whitesides \& Grzybowski(2002)Whitesides and
  Grzybowski]{whitesides_self-assembly_2002}
Whitesides, G. and Grzybowski, B.
\newblock {Self-Assembly} at all scales.
\newblock \emph{Science}, 295:\penalty0 2418, 2002.

\bibitem[Winfree et~al.(1988)Winfree, Liu, Wenzler, and
  Seeman]{winfree_design_1988}
Winfree, E., Liu, F., Wenzler, L., and Seeman, N.
\newblock Design and {Self-Assembly} of two-dimensional {DNA} crystals.
\newblock \emph{Nature}, \penalty0 (394):\penalty0 539--544, 1988.

\bibitem[Wright(1932)]{wr32}
Wright, S.
\newblock The roles of mutation, inbreeding, crossbreeding and selection in
  evolution.
\newblock \emph{Proceedings of the XI International Congress of Genetics},
  8:\penalty0 209--222, 1932.

\bibitem[Wu(2012)]{wu_oxford_2012}
Wu, N.
\newblock Oxford haskell user group, 2012.

\end{thebibliography}
\bibliographystyle{icml2021}

\appendix
\onecolumn

\section{Classification of $S^{32}_{3,8}$}
\label{app:s38}

\begin{figure}
    \centering
    \includegraphics{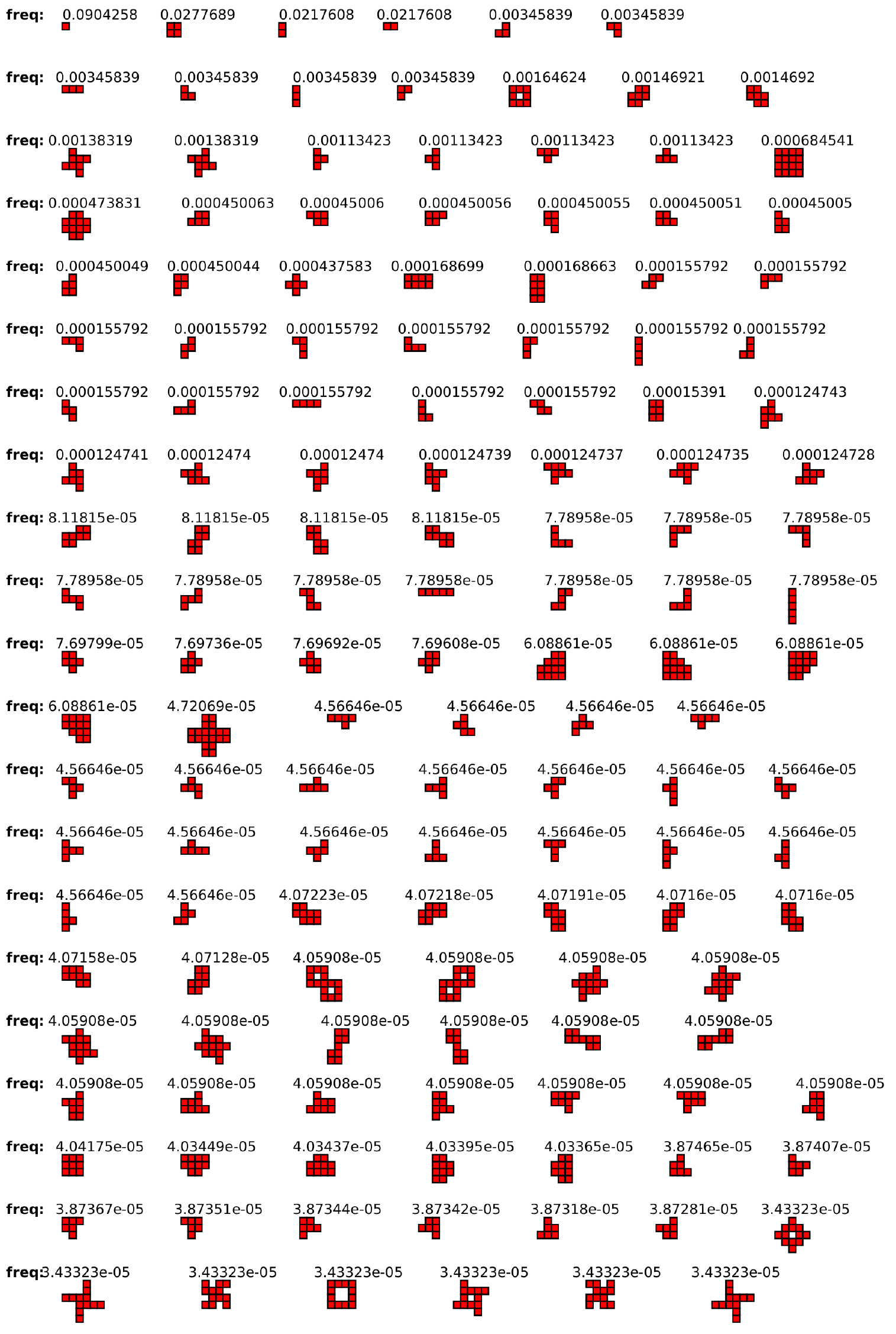}
    \caption{[1] Classification of $S^{32}_{3,8}$  at $7$ redundancy assemblies for detection of steric nondeterminism, ordered by frequency. Frequencies are calculated with respect to the whole genome search space of size $2^{32}$.}
    \label{fig:classification_S32_3_8__page1}
\end{figure}

\begin{figure}
    \centering
    \includegraphics{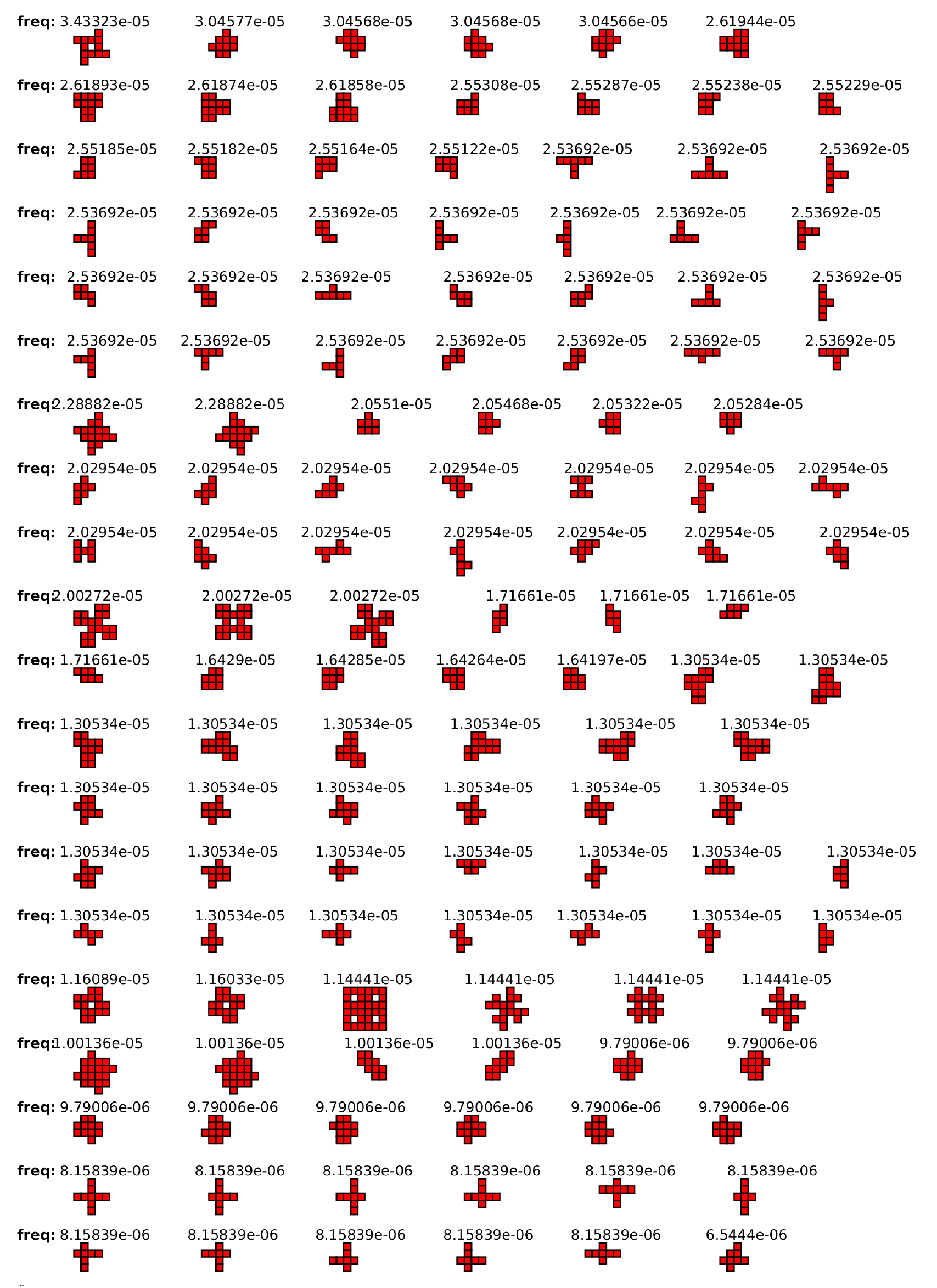}
    \caption{[2] Classification of $S^{32}_{3,8}$ at $7$ redundancy assemblies for detection of steric nondeterminism, ordered by frequency. Frequencies are calculated with respect to the whole genome search space of size $2^{32}$.}
    \label{fig:classification_S32_3_8__page2}
\end{figure}

\begin{figure}
    \centering
    \includegraphics{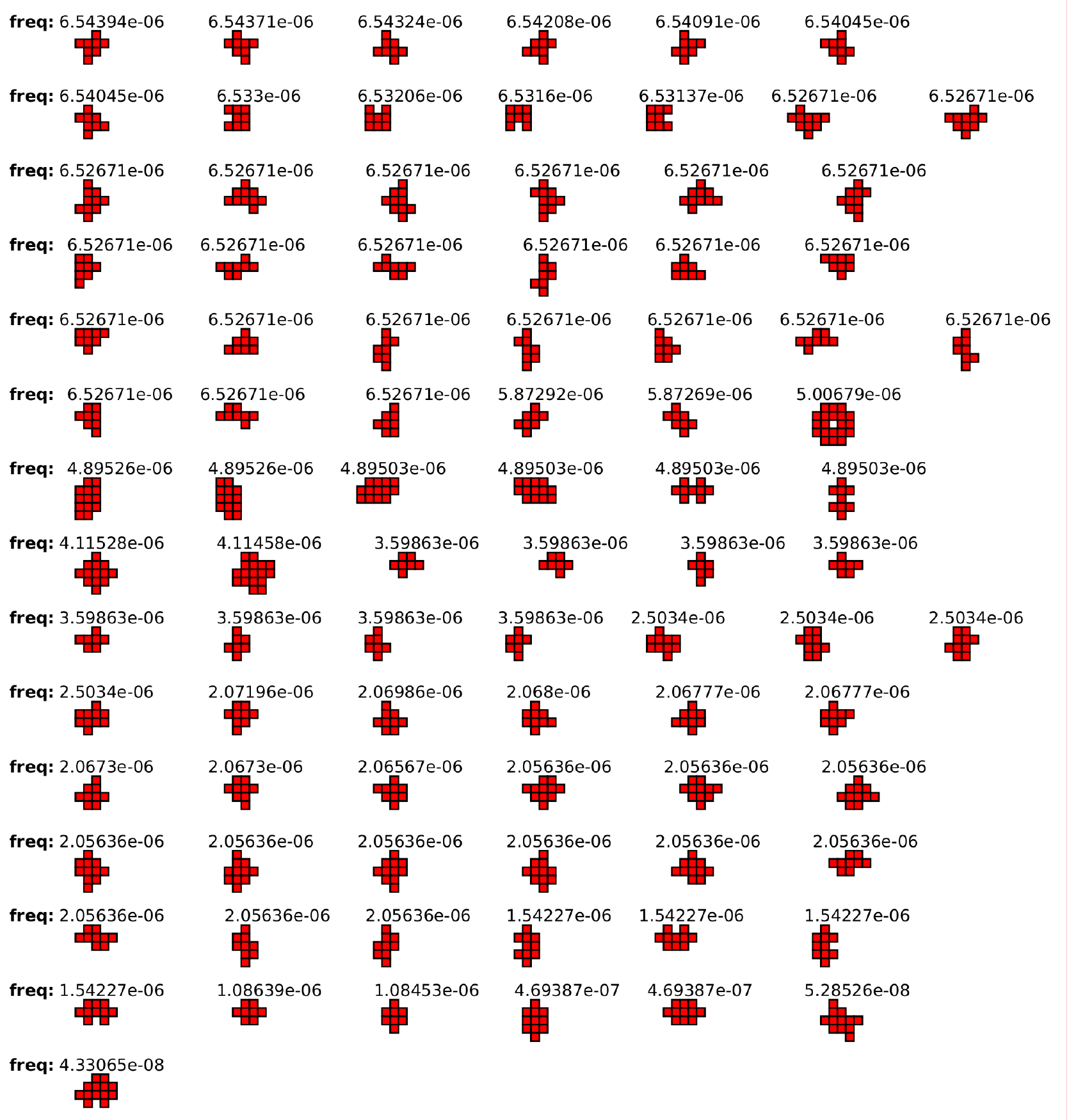}
    \caption{[3] Classification of $S^{32}_{3,8}$ at $7$ redundancy assemblies for detection of steric nondeterminism, ordered by frequency. Frequencies are calculated with respect to the whole genome search space of size $2^{32}$.}
    \label{fig:classification_S32_3_8__page3}
\end{figure}

\begin{figure}
    \centering
    \includegraphics{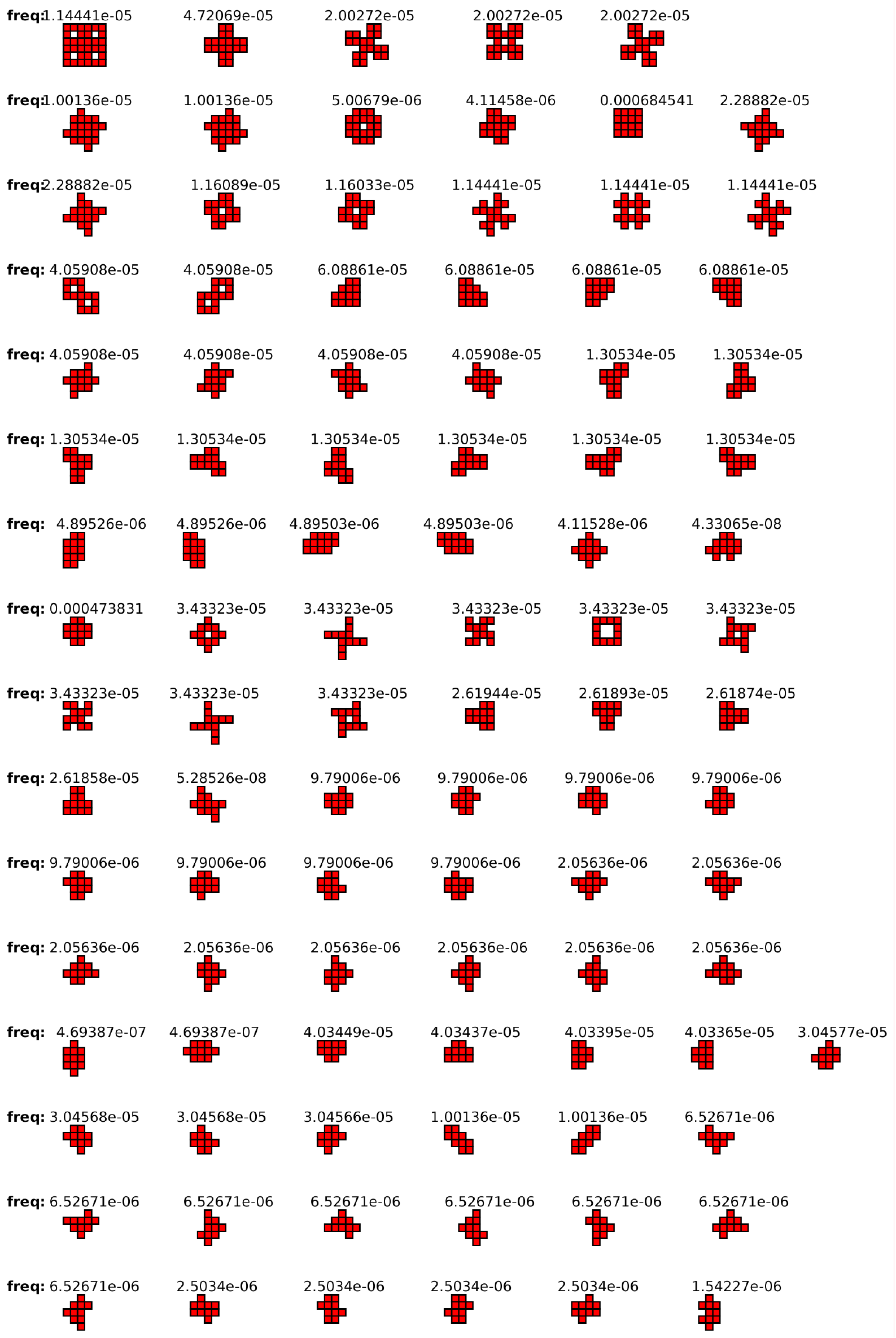}
    \caption{[4] Classification of $S^{32}_{3,8}$   at $7$ redundancy assemblies for detection of steric nondeterminism, ordered by frequency. Frequencies are calculated with respect to the whole genome search space of size $2^{32}$.}
    \label{fig:classification_S32_3_8__page4}
\end{figure}

\begin{figure}
    \centering
    \includegraphics{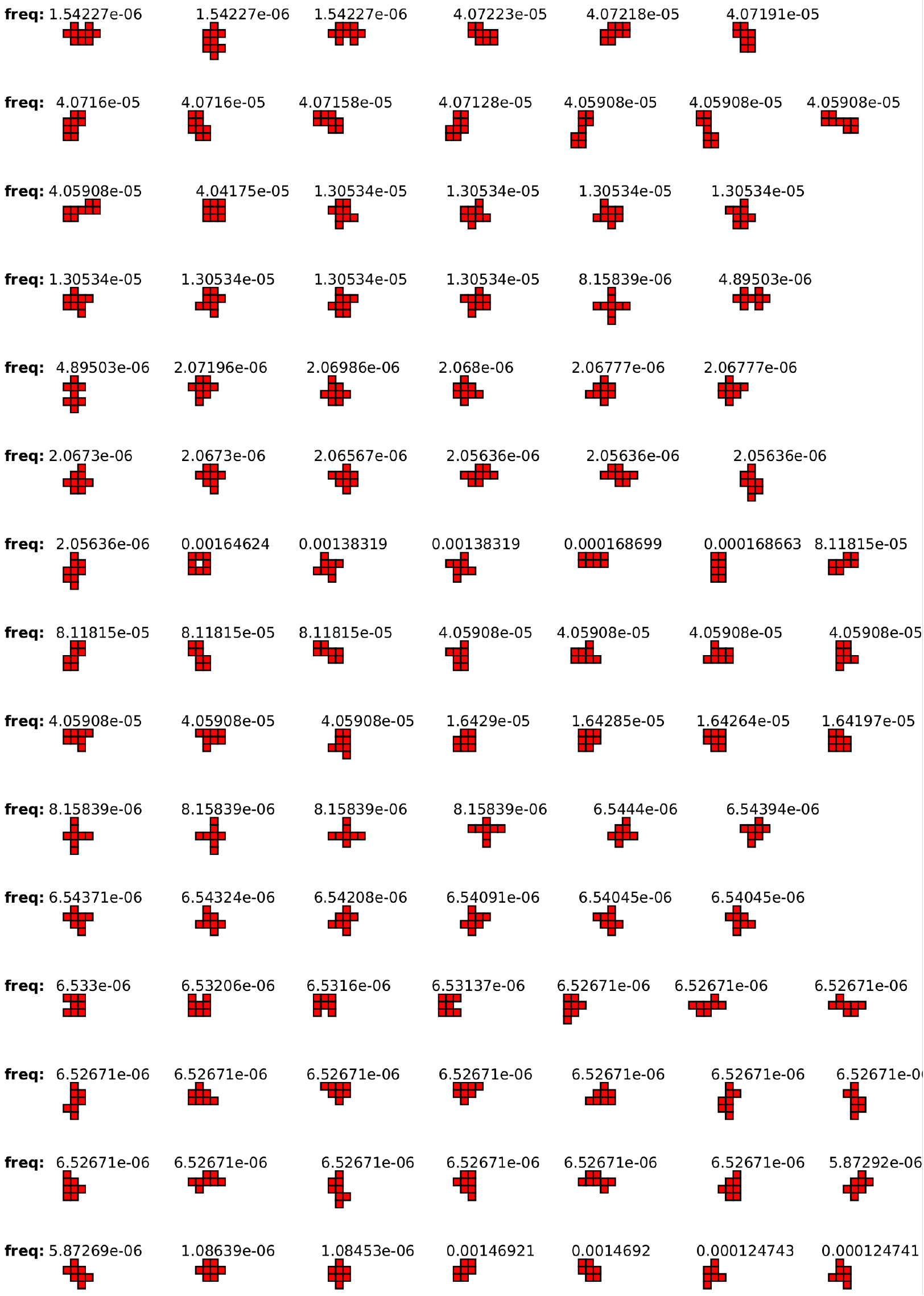}
    \caption{[5] Classification of $S^{32}_{3,8}$  at $7$ redundancy assemblies for detection of steric nondeterminism, ordered by frequency. Frequencies are calculated with respect to the whole genome search space of size $2^{32}$.}
    \label{fig:classification_S32_3_8__page5}
\end{figure}

\begin{figure}
    \centering
    \includegraphics{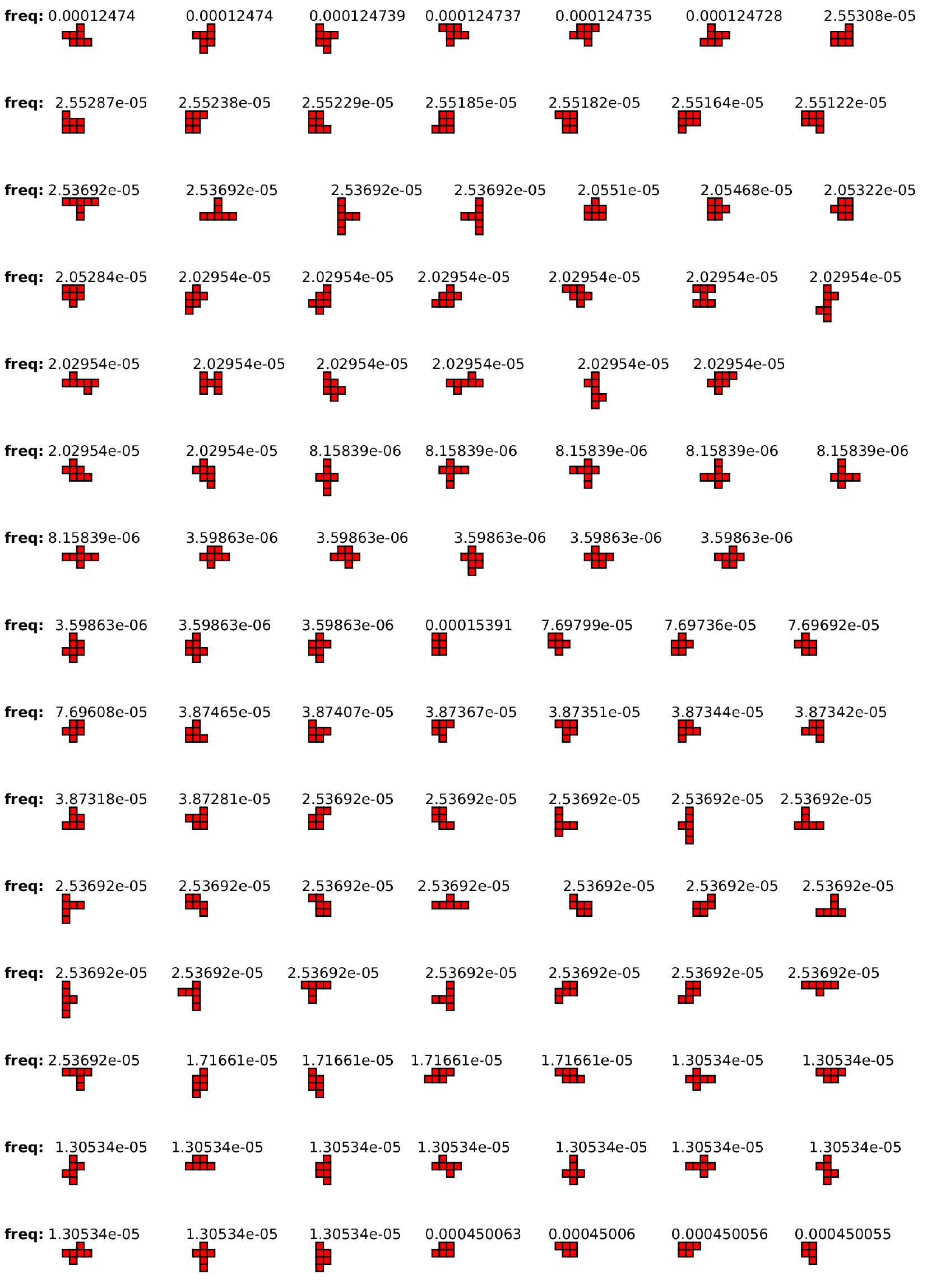}
    \caption{[6] Classification of $S^{32}_{3,8}$ at $7$ redundancy assemblies for detection of steric nondeterminism, ordered by frequency. Frequencies are calculated with respect to the whole genome search space of size $2^{32}$.}
    \label{fig:classification_S32_3_8__page6}
\end{figure}

\begin{figure}
    \centering
    \includegraphics{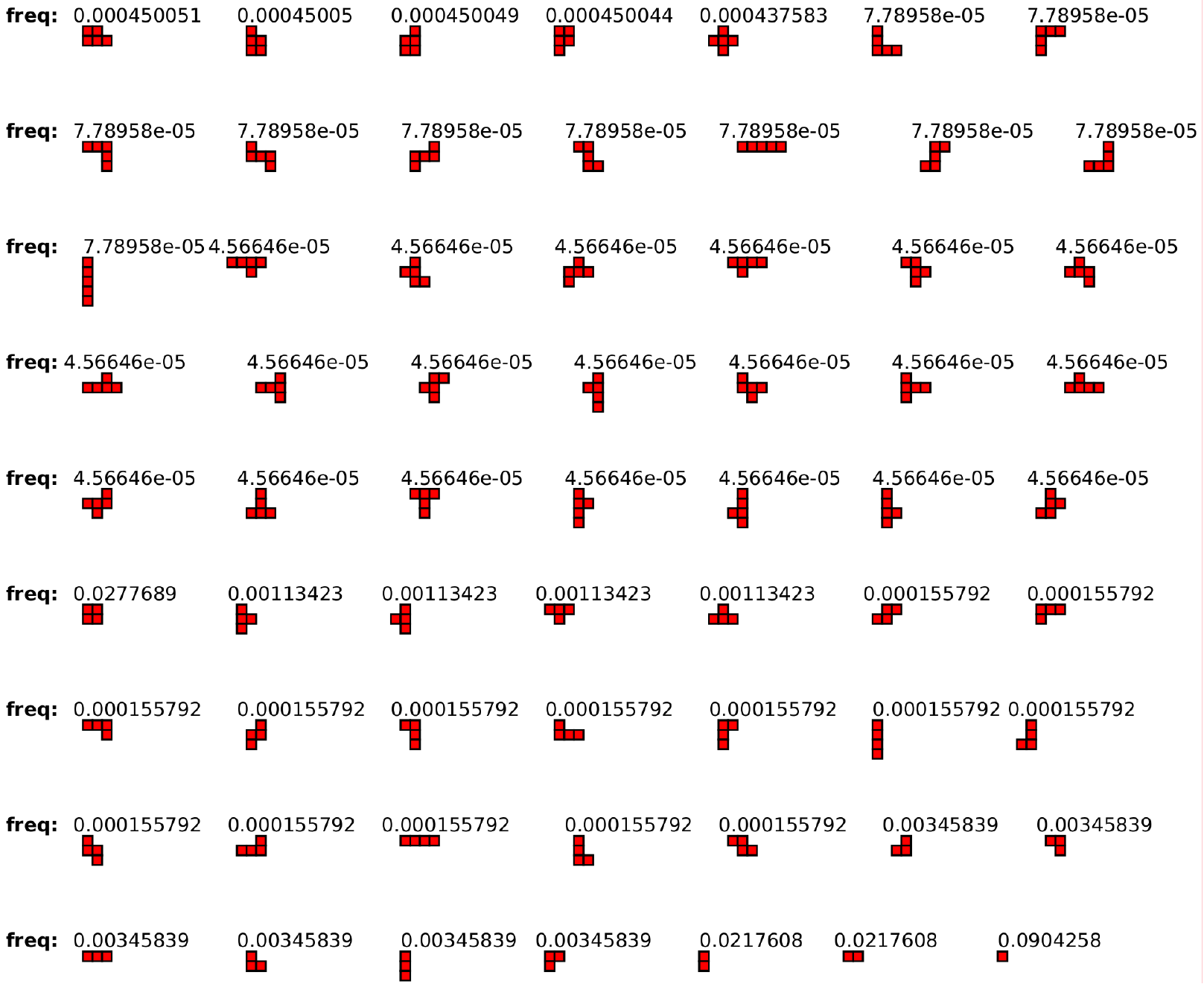}
    \caption{[7] Classification of $S^{32}_{3,8}$ at $7$ redundancy assemblies for detection of steric nondeterminism, ordered by frequency. Frequencies are calculated with respect to the whole genome search space of size $2^{32}$.}
    \label{fig:classification_S32_3_8__page7}
\end{figure}

See Figures \ref{fig:classification_S32_3_8__page1}, \ref{fig:classification_S32_3_8__page2}, \ref{fig:classification_S32_3_8__page3},
\ref{fig:classification_S32_3_8__page4},
\ref{fig:classification_S32_3_8__page5},
\ref{fig:classification_S32_3_8__page6}, and
\ref{fig:classification_S32_3_8__page7}.

\section{Formalistic treatment of self-assembly}

\subsection{Winfree’s abstract tile assembly model}
\label{app:winfree}

Winfree’s model extends planar self-assembly of Wang tiles \citep{rothemund_program-size_2000}. A Wang tile T is a unit
square with edges colored from a binding domain $\Sigma$ \citep{wang_proving_1961}. The binding domain $\Sigma$ includes
the neutral color null, which does not bond to any other color. A set of symmetric strength
functions $g$ indicates the interaction strengths between any two colors. Matching colors are
assigned some positive integer value interaction strength, all other combinations are set to
zero. Tiles may not be rotated nor flipped and there exists an empty tile which does not bond
to any other tile.
A \textit{tileset} is the set of all tile types that may be used in the assembly process. Each tile type
may be used arbitrarily often. A particular tile occupation pattern of the assembly plane is
called a configuration. An individual step in the assembly process consists of randomly adding
a further bonding tile from the tileset to the existing configuration.
The free energy $G(C)$ of a configuration $C$ is defined to be the sum over all interaction
strengths between the assembled tiles \citep{rothemund_program-size_2000}. The relative sizes of the free energy of a configuration and an abstract temperature $\mathcal{T}$ determine whether a given assembly is temperature-stable, i.e. that it is fully connected and will not tend to dissociate into two or more smaller
configurations over time. In general, if a tile $T_i$
is added to an existing $\mathcal{T}$-stable configuration
A, the resulting configuration B is $\mathcal{T}$-stable iff the sum over all bonding strengths to neighbour tiles of $T_i$ is greater or equal to $\mathcal{T}$ \citep{rothemund_program-size_2000}.
In order to fully specify a given assembly process, we need to specify a tile system $\mathbb{T} =\left\langle T, S, g, \mathcal{T}\right\rangle$. Here, T is a tileset including the \textit{empty} tile and \textit{Seed} is a set of \textit{seed tiles} at
specified positions, i.e. the initial configuration consists of Seed and empty tiles everywhere
else.
Each \textit{tile system} $\mathbb{T}$ has a map Prod:$\mathbb{T}\rightarrow$ \textit{assembly configurations}. The set Term($\mathbb{T}$) is the
subset of Prod($\mathbb{T}$) which includes all configurations that \textit{terminate} in finite time. If every possible assembly path through Prod($\mathbb{T}$) results in an element of Term($\mathbb{T}$) in finite time and the cardinality of Term($\mathbb{T}$) is finite, then $\mathbb{T}$ is said to be \textit{halting}. \citep{rothemund_program-size_2000}

\subsection{Modifications to Winfree’s aTAM}
\label{app:mod}

Due to its two-dimensional nature, it has been suggested that abstract tile assembly model
(aTAM) may be used to approximate self-assembly of proteins or protein sub-units on the
surface of a planar substrate \citep[Fig. 1]{johnston_evolutionary_2011}. In this light, \citet{johnston_evolutionary_2011} have suggested a
modified version of aTAM, which I refer to as \textit{JaTAM}.
I now extend Winfree's \textit{aTAM} formalism to Johnston's model.
JaTAM fixes $\mathcal{T}$ to the value of $1$. This makes sense if building block interaction strengths
are much larger than thermodynamic noise. The restriction $\mathcal{T}=1$ means that for any
configuration $C$, $G(C)$ is sufficiently large such that $C$ is $\mathcal{T}$-stable. So $g$ can be characterised by a binary matrix $\mathcal{M}$. \textit{JaTAM} fixes $\mathcal{M}$ to $M_{ij}=(1-i\mod2)\delta_{i(j+1)}+(i\mod2)\delta_{i(j-1)}$, i.e. colors bond in adjacent pairs, which is in contrast to
the color-matching scheme of \textit{aTAM}. Adjacent pair-bonding may mimick the bonding scheme between complementary DNA bases. This \textit{heterogeneous strength function} produces different
phenomenology from \textit{aTAM} \citep{ahnert_self-assembly_2009}.
\citet{johnston_evolutionary_2011} allow orthogonal rotations of Wang tiles, such that each configuration additionally contains orientation flags of all assembled tiles. \textit{Seed} is defined to be the first element
in the tile set $T$, which is positioned at the center of an empty grid.
Two configurations are equivalent iff all assembled tiles at each point of the assembly grid are equivalent in type and orientation. If the positions of all empty tiles coincide between
two configurations, then they are \textit{shape-equivalent} \citep{johnston_exploration_2010}. A given $\mathbb{T}$ is deterministic iff
all assembly paths through $Prod(\mathbb{T})$ terminate in equivalent configurations. Otherwise, $\mathbb{T}$ is
\textit{non-deterministic}. Iff a given configuration $A$ contains non-empty tiles at positions outside
an origin-centred square of dimension $d$, then $A$ is said to be unbounded by $d$. Otherwise, $A$ is
bounded by $d$. Iff $\mathbb{T}$ is halting, then we can always find a finite, positive $z$ such that $Prod(\mathbb{T})$
is bounded by $z$.
Let the \textit{assembly frontier} be the set of all assembly grid positions that are occupied by an
empty tile and have at least one non-empty neighbour tile.
Let's choose some \textit{non-deterministic} $\mathbb{T}^*$
. For any configuration $A \in Prod(\mathbb{T}^*)$, let $B$ denote
the set of configurations generated after one further tile $\alpha$ has been assembled on the assembly
frontier of $A$. Let $P$ be the point where tile $\alpha$ has been assembled.
If $B$ contains a subset of configurations which correspond to equal $P$ but are not all equivalent
to each other, then $\mathbb{T}^*$
is said to exhibit trivial non-determinism.
\textit{Steric non-determinism} arises if at least two partial assembly paths exclude each other without
the involvement of trivial non-determinism. Phenomenologically, this arises when two or more
independently growing "arms" converge \citep{johnston_exploration_2010}.
$\mathbb{T}^*$ may exhibit trivial non-determinism and steric non-determinism at the same time.
I now proceed by extending Johnston’s terminology by defining a shape-similarity measure. I
define the shape-difference $\text{shapediff}(A, B) \in \mathbb{N}$ of two configurations $A, B$ to be the number
of non-coinciding empty tile positions between $A$ and $B$. Then, iff $A, B$ are just bounded
by $d$, the shape-similarity $\text{shapesim}(A, B) \in [0, 1]$ of $A, B$ is defined to be equivalent to
$1-\text{shapediff}(A, B)/d^2$.
As tile-shaped building blocks may bind to a substrate from either above or below, and
\citet{johnston_evolutionary_2011} suggest a future assembly-system, which allows tiles bonding from two sides
to be arbitrarily flipped in the substrate plane. In this paper, however, I will restrict myself
to JaTAM, which is reasonable if we assume that tiles only bind to the substrate from
a single side.

\subsection{Encoding genome information}
\label{app:enc}

By Shannon \citep{shannon_mathematical_1948}, in order to label $|\mathfrak{G}_D|$ \textit{genome microstates}, an average number of $log_2
|\mathfrak{G}_D|$
bits are needed. While, in principle, we are free to choose any such encoding, there are particular choices of binary encodings that are more natural than others based on characteristics
of the evolutionary operators acting on the \textit{search space} $S$. These characteristics should be
equivalent to their counterparts in natural DNA encoding.

Applying a suitable mutation operator $\mathfrak{M}$ to some genome $G\in S$ needs to yield another
element $\in S$. A formal way of stating this is to say that $S$ is closed under the operator $\mathfrak{M}$.
Furthermore, any $X\in S$ must be transformable into any arbitrary element $\in S$ with non-zero
probability in a finite number of steps, i.e. $\mathfrak{M}$ is \textit{permuting}.
A suitable single-point crossover operator $\mathfrak{C}$ needs to be such that $S$ is closed under it as well.
However, it is not a permuting operator, which can be seen by choosing two genomes with
uniform base occupations - however often we recombine the two, we will never be able to
create a permutation that contains the third or fourth base type used by DNA encoding. A
characteristic property of $\mathfrak{C}$ is, however, that it preserves the relative order of all genes involved.

I conclude that with respect to mutation, the choice of what bases constitute a particular gene
is arbitrary. For \textit{single-point crossover}, however, relative order of genes has to be preserved
and this can in the general case only be defined if all bases belonging to a particular gene sit
at adjacent locations in the genome.
The observation of genetic drift, however, strongly depends on the correlation between a
particular gene and its phenototypic expression. Whether it is possible to define a gene-interpretation for Johnston’s model that allows for strong correlations between genotypes and
phenotypes is a key question for further research.
\textit{Kolmogorov complexity} $\mathcal{K}(s_b)$ is defined as the shortest program size written in a language $\mathbb{L}$
which can output a bitstring $s_b$. We will in the following define $\mathbb{L}$ to be the language of a
\textit{Universal Turing Machine}. A phenotype shape in aTAM can be defined by supplying a list of
the locations of all non-empty tiles, which I call a list representation of that shape. \citet{winfree_design_1988} showed that $\mathcal{K}$ of the list representation of a scale-free phenotype shape is proportional to the
minimum number $n$ of tile types, with respect to some tile system $\mathbb{T}$, that \textit{deterministically}
assembles that shape  \citep{soloveichik_complexity_2007}. Even though the relationship to $\mathcal{K}$ of heterogeneous strength
functions is not strictly proven \citep{soloveichik_complexity_2007}, \citet{ahnert_self-assembly_2009} provide an algorithm which computes
n with respect to a phenotype shape assembled by JaTAM and is of the same order as the
respective Kolmogorov complexity\citep{ahnert_self-assembly_2009}.
With respect to JaTAM, I conclude that a strongly phenotype-correlated gene-interpretation
might consist of gene $1$ being the minimum number of tile types required to compute the
phenotype shape encoded in the genome and a gene $2$ which encodes a notion of \text{shapescale}.
Such an encoding does not exclude a constant bitstring binary encoding with respect to $M$.
Defining the action of the operator $\mathfrak{C}$ on such a binary encoding, however, is non-trivial and
subject to further research.

I finally describe the constant bitstring binary encoding used by \citet{johnston_evolutionary_2011} and how the respective action of $\mathfrak{M}$ may be defined. Johnston encodes a general tile set in the genome by
sequentially encoding each tile in the tile set to a bitstring, starting with the seed tile. This
is achieved by subsequently encoding all four colors of the tile in the order north-east-southwest, as looking down on the configuration grid. The number of bits required to encode a
particular color is $log_2 \lVert\Sigma\rVert$ and Johnston always minimises the number of bits required.
The action of $\mathfrak{M}$ on this representation of $S$ is to flip a number $n\mu$ of distinct bits, where $n\mu$ is Poisson-distributed with mean $\mu$. As this implies a non-zero probability for the flipping of
just a single bit, sequential $\mathfrak{M}$-operations can be used to traverse arbitrary Hamming weight
distances in search space and $\mathfrak{M}$ is thus permuting. To achieve closure under $\mathfrak{M}$, I conclude
that Johnston’s encoding restricts $\lVert\Sigma\rVert$ to integer powers of $2$. It is now trivial to define the
map tileset, which decodes Johnston’s binary encoding into a tileset.

\section{On Flow Charts, Unit Tests and UML diagrams}
\label{app:uml}

In many situations, flow charts are an insightful way to represent the procedural large-scale
flow of a program. In order to turn a piece of code into a flow chart, one first represents
all branches and loops as an empty flow-chart and then fills in the pictograms with some
short-hand summary of the action of the underlying code. The content of each pictogram
then defines a particular code unit.
The functionality of each code unit may be described by the help of abstract pseudo code. In
order to ensure the integrity of a code unit, a carefully-written unit test should be designed.
A unit test defines a set of input values to a code unit and specifies the desired set of output
values. The distribution of input values can be a random distribution or taylored to test particularly suspicious parameter ranges. The assertion strength of a unit test is thus dependent
on the choice of input values.
A principal advantage of unit tests is that they can act on black boxes, i.e. program parts whose
precise internals that cannot or should not be presented to a general audience. By publishing
the input parameters of the unit test, however, and the resulting output parameters, one can
display code integrity without the need of detailed code disclosure or analysis. Furthermore,
large and complex code units may be broken down into simpler, independent components
which can be unit-tested individually.
If a piece of software is designed in an object-oriented way, flow charts may be replaced by
Unified-Modeling language (UML) diagrams. UML is a very broad standard allowing the
graphical representation of flow processes in a large number of domains. A particular type
of UML diagram is a class diagram which represents all classes, their attributes and their
relationships with each other. There exist tools that can generate UML class diagrams from
source code and also some that allow the generation of source code from such a diagram.

\subsection{Tournament Selection in Haskell}

In order to illustrate how functional programming is independent of program flow, I attach
snippets in \textbf{C} and \textbf{Haskell} used for tournament selection: We supply a list of weights and a cut-off index (which is usually chosen randomly) and would like our function to return the counter index of the element where the sum of the weights is greater or equal to the cut-off index.

\subsubsection{C-implementation}

\begin{minted}{c}
unsigned int tourn(void){
    for(int i=0;i<mWeightLen;i++){
        partsum += weights[i];
        if(partsum >= cutoff){
            counter = i;
            break;
        }
    }
    return counter;
}
\end{minted}

We initialise \textit{global} parameters as follows:
\begin{minted}{c}
#define mWeightLen 4
float weights[mWeightLen] = {2.0,3.0,4.0,1.0}
unsigned int cutoff = 10;
unsigned int counter = 0;
float partsum = 0.0f;
\end{minted}

\subsubsection{Haskell-implementation}

\begin{minted}{haskell}
tourn :: [Double] -> [Integer] -> Double -> Double -> Integer
tourn (x:xs) (y:ys) z p = if (p+x) >= z then y else tourn xs ys z (p+x)
\end{minted}

The parameters should be initialised as follows:

\begin{minted}{haskell}
weights = [2.0,3.0,4.0,1.0]
indices = [0.. fromIntegral (length weights)]
cutoff = 10.0
counter = 0.0
\end{minted}

We note that \textit{tourn} is simply a \textit{curried}, \textit{recursive} function. The internal flow of this function is determined by the \textit{compiler} only.

\section{ Efficient bit-wise implementations of common evolutionary operators}

\subsection{Mutation from Poisson Distribution}

An implementation of mutation from Poisson distribution follows below. Note that we could
reduce the required storage space for the bitmask, of course at the expense increasing computing time, by using a dynamic list to keep track of the flipped bits. Dynamic lists in \textit{shared memory}, however cannot be efficiently implemented as there is no \textit{heap} available.

\begin{minted}{c}
__forceinline__ __device__ void ga_MutationSophisticated( unsigned char (&s_ucGenome)
                                        [m_ga_NR_THREADS_PER_BLOCK][mByteLengthGenome],
                                        unsigned char (&s_ucBufGenome)
                                        [m_ga_NR_THREADS_PER_BLOCK][mByteLengthGenome],
                                        float mutation_rate,
                                        curandState *state){
    for(int i=0;i<mByteLengthGenome;i++){
        s_ucBufGenome[mTHREAD_ID][i] = 0;
    }
    float r_fNrMutations = ga_uiPoissonDistribution(mutation_rate, state);
    if(r_fNrMutations > mBitLengthGenome) r_fNrMutations = mBitLengthGenome;
    float r_fRandBuf;
    short r_ssBitOffset, r_ssByteOffset;
    bool r_bRetry = false;
    for(int i=0;i<r_fNrMutations;i++){
        r_fRandBuf = curand_uniform(state) * mBitLengthGenome;
        r_ssBitOffset = (signed short) r_fRandBuf % 8 ;
        r_ssByteOffset = ((signed short) r_fRandBuf - r_ssBitOffset) / 8;
        if( ! ( s_ucBufGenome[mTHREAD_ID][r_ssByteOffset] & (1 << r_ssBitOffset))){
            s_ucBufGenome[mTHREAD_ID][r_ssByteOffset] += (1 << r_ssBitOffset);
            r_bRetry=false;
        } else {
            i--;//Try again to mutate!
        }
    }
    for(int i=0;i<mByteLengthGenome;i++){
        s_ucGenome[mTHREAD_ID][i] = mXOR(s_ucGenome[mTHREAD_ID][i],
        s_ucBufGenome[mTHREAD_ID][i]);
    }
}
\end{minted}

\subsection{Single-point crossover}
\label{app:reinterpret}

An efficient implementation of \textit{single-point crossover}, which uses byte-wise operations instead
of bit-wise operations. A further performance gain would be achieved by further increasing
the granularity using \textit{reinterpret\_cast} as the \textit{genome bytelength} exceeds a couple of bytes (up to $8$ for double precision).

\begin{minted}{c}
__forceinline__ __device__ void ga_CrossoverSinglePoint(unsigned char (&s_ucGenome)
                                                            [m_ga_NR_THREADS_PER_BLOCK]
                                                            [mByteLengthGenome],
                                                        unsigned char *g_ucGenomes,
                                                        unsigned int r_uiCutoffIndex,
                                                        curandState *state){
    unsigned short int r_uiCrossoverPoint = curand_uniform(state) * mBitLengthGenome;
    unsigned short int r_uiCrossoverBitOffset = r_uiCrossoverPoint % 8;
    unsigned short int r_uiCrossoverByte = 
        (r_uiCrossoverPoint - r_uiCrossoverBitOffset) / 8;
    unsigned int r_uiIndexBuffer;
    for(int j=0;j<=r_uiCrossoverByte;j++){
        if(j == r_uiCrossoverByte){
            s_ucGenome[mTHREAD_ID][j] = ( s_ucGenome[mTHREAD_ID][j]
                          & ( 0xFF >> r_uiCrossoverBitOffset ) )
                          + ( g_ucGenomes[r_uiCutoffIndex * mByteLengthGenome + j]
                          & ( 0xFF << ( 8 - r_uiCrossoverBitOffset ) ) );
        } else {
            s_ucGenome[mTHREAD_ID][j] = 
                g_ucGenomes[r_uiCutoffIndex * mByteLengthGenome + j];
        }
    }
}
\end{minted}

\subsection{Uniform crossover}

An efficient implementation of uniform crossover, where both genomes are randomly interleaved on bit-level. Note that interleaving occurs in multiples of $4$ bytes, which is the number
of bytes that a call to the random number function generates.

\begin{minted}{c}
__forceinline__ __device__ void ga_CrossoverUniform(unsigned char (&s_ucGenome)
                                                        [m_ga_NR_THREADS_PER_BLOCK]
                                                        [mByteLengthGenome],
                                                    unsigned char *g_ucGenomes,
                                                    unsigned int r_uiCutoffIndex,
                                                    curandState *state ){
    unsigned int r_uiNumberOfRandCallsRequired = (unsigned int)
                ((mByteLengthGenome - mByteLengthGenome%sizeof(float))/sizeof(float) + 1);
    float r_fRandBuffer;
    unsigned char r_ucRandMask;
    unsigned int r_uiIndexBuffer;
    for(int i=0;i<r_uiNumberOfRandCallsRequired;i++){
        r_fRandBuffer = curand_uniform(state);
        for(int j=0;j<sizeof(float);j++){
            r_ucRandMask = (unsigned int) ((reinterpret_cast<int&>(r_fRandBuffer)
                            & (0xFF << 8 * j)) >> 8 * j);
            r_uiIndexBuffer = i*sizeof(float) + j;
            if(r_uiIndexBuffer < mByteLengthGenome){
                s_ucGenome[mTHREAD_ID][r_uiIndexBuffer] = (s_ucGenome[mTHREAD_ID]
                                [r_uiIndexBuffer] & r_ucRandMask) + (g_ucGenomes
                                [ga_xGlobalAnchor(r_uiCutoffIndex * mByteLengthGenome +
                                r_uiIndexBuffer)] & (~r_ucRandMask)) ;
            }
        }
    }
}
\end{minted}

\section{PyCUDA - tips and tricks}

This section includes some work-arounds which I needed to employ in order to get things work
and which I could not find in any forum or document.

\subsection{Getting \textit{curand} to work}
\label{app:mangling}

\textit{PyCUDA} is designed such that python code can call kernel functions defined in \textit{C/C++}.
For this purpose, it requires knowledge of the exact reference name of the function used by
the compiler. In \textit{C}, this is merely the name of the function as defined by the programmer. \textit{C++}, however, supports functions of equal naming but different \textit{namespaces}. In order to keep track of these functions, the compiler \textit{name-mangles} them to some label unpredictable to \textit{PyCUDA}. As \textit{curand\_kernel}-header uses different \textit{namespaces}, it requires name-mangling
to be activated, while the rest of the kernel code may not feature it.
This problem may be solved by setting the Source Module-Option `no\_extern\_c' to `True' and
enclosing everything but the header in \textit{extern} `C'-brackets.

\subsection{Unrolling for-loops using meta-templating}
\label{app:unroll}

Interestingly, unrolling small for-loops may result in performance gains of up to a factor of 2, 
as has been observed previously \cite{murthy_optimal_2010}. An automatised way of unrolling loops at runtime is the
usage of a metatemplating language, such as \textit{Jinja2}\cite{ronacher_jinja2_2012}. The following code segment demonstrates the $A$-times unrolling of a loop:
\begin{minted}{c}
{% for idx in range(A) %}
    this->data.flags[__xThreadInfo.BankId()].set_Red({{ idx }});
    this->Assemble_Movelist(__xThreadInfo, __xGenomeSet);
    __syncthreads();
{% endfor %}
\end{minted}

\section{Selected Algorithms}

\subsection{Bit operations and casting}

Using genome bit-representation instead of e.g. \textbf{boolean} representation allows the usage of very fast bit-operations which directly translate into x$86$ processor instructions. For reference, I include basic usage of bit operations on the nth bit in \textit{C/C++}:

\paragraph{Bit setting}
\begin{minted}{c}
genome |= (1 << n);
\end{minted}

\paragraph{Bit clearing}
\begin{minted}{c}
genome &= ~(1 << n);
\end{minted}

\paragraph{Bit testing}
\begin{minted}{c}
genome & (1 << n);
\end{minted}

\paragraph{Bit toggling}
\begin{minted}{c}
genome ^= (1 << n);
\end{minted}

\paragraph{Modulo 2 operations}
\begin{minted}{c}
n%2 == n & 0x1;
\end{minted}

All these operations can be effectively parallelised for a larger number of bits using \textit{bitmasks}.

\subsection{The use of reinterpret\_cast}

The data type of a memory segment may be re-interpreted as a different data type by the
usage of either \textit{unions} in \textit{C} and / or \textit{reinterpret\_cast} in \textit{C++}. With caution, pointers reinterpreted by \textit{reinterpret\_cast} may be de-referenced. In this way, char-arrays may be initialised with a step length of up to $16$-byte on GPU, which I found to be about $4$ times faster than char-wise initialisation.
For \textit{gcc}-based compilers (such as likely \textit{nvcc}), dereferencing a reinterpreted pointer may trigger the following warning: warning: type punned pointer will break strict-aliasing rules. This warning can be eliminated by setting the flag \textit{-fno-strict-aliasing}. This does take away some
optimisation options from the compiler, however, from my experience is usually acceptable
performance-wise.

\subsection{Bit operations and casting in Python}

As python does not natively implement fixed-size numeric datatypes, the usage of bit operations may not be as performant as for \textit{C/C++}. However, equivalent operators may be used
e.g. for calculating \textit{Jenkin’s hash} \cite{jenkins_hash_1997}. The size of the data type may then be selected aftwards using a bitmask (i.e. to select a $32$-bit data type from a pythonic number x, we may use $\text{x\&=0xffffffff}$ to set higher bits to zero.)
As python does not offer pointer manipulation natively, one needs to employ a foreign-function
interface such as \textit{ctypes} in order to implement \textit{C/C++}-style union data types and/or type casting.

\subsection{One-at-a-time Hash}

The following is an efficient implementation of a 32-bit hash function taking one byte at a
time by Jenkins \cite{jenkins_hash_1997}. This so called One-at-a-time hash has a decent avalanche behaviour and
is very fast.

\begin{minted}{c}
__device__ void jenkins_init(int &hash){
    hash = 0;
}
__device__ void jenkins_add(char key, int &hash){
    unsigned int tmphash=hash;
    tmphash += key;
    tmphash += (tmphash << 10);
    tmphash ^= (tmphash >> 6);
    hash=tmphash;
}
__device__ unsigned int jenkins_clean_up(int &hash){
    unsigned int tmphash=hash;
    tmphash += (tmphash << 3);
    tmphash ^= (tmphash >> 11);
    tmphash += (tmphash << 15);
    hash = tmphash;
    return hash;
}
\end{minted}

\subsection{Reentrant LIFO-Stack}
\label{app:lifo}

I thank \textit{tera} \citep{schmielau_implementing_2012} for making me aware of undercurrent and overcurrent detection issues arising from the use of atomic operations in \textit{CUDA} with unsigned data types. The following is a push-function, designed by \textit{tera}, which has been made multi-threading safe for use in a multi-threaded \textit{LIFO}-stack implementation:

\begin{minted}{c}
__device__ bool x_push(int entry, unsigned int *pos_p, int max_length, int* stor){
    unsigned int BankID = threadIdx.x % 32;
    unsigned int pos = atomicAdd(pos_p, 1);
    if(pos < max_length){
        stor[BankID + pos*32] = entry;
        return true;
    } else {
        atomicSub(pos_p, 1);
        return false;
    }
}
\end{minted}

Note: \textit{pos\_p} should be initialised to $0$. Note also that changes in the usage of atomic functions which would be valid for \textit{boost} threads may readily lead to data inconsistencies for usage with \textit{CUDA}.

\subsection{Parallelised grid cropping}
\label{app:cropping}

The following parallelised code determines the upper left and the lower right crop corner of
an assembly grid:

\begin{minted}{c}
if(__xThreadInfo.WarpId()==0){
    this->data.corner_lower[__xThreadInfo.BankId()]=make_int2(0,0);
    this->data.corner_upper[__xThreadInfo.BankId()]=
        make_int2(m_fit_DimGridX-1, m_fit_DimGridY-1);
    this->data.assembly_size[__xThreadInfo.BankId()]=0;
}
__syncthreads();
short offset = (m_fit_DimGridX*m_fit_DimGridY) % mBankSize;
short myshare = (m_fit_DimGridX*m_fit_DimGridY - offset) / mBankSize;
int off_x=0, off_y=0;
for(int i=0;i<myshare;i++){
    off_x = (myshare*__xThreadInfo.WarpId()+i) % m_fit_DimGridX;
    off_y = (myshare*__xThreadInfo.WarpId()+i-off_x) / m_fit_DimGridX;
    if(this->data.grid.data.multi_d[off_x][off_y][this->data.flags
        [__xThreadInfo.BankId()].get_ucRed()]
        [__xThreadInfo.BankId()].get_xCell()!=mEMPTY_CELL){
            atomicMin(&this->data.corner_upper[__xThreadInfo.BankId()].x, off_x);
            atomicMin(&this->data.corner_upper[__xThreadInfo.BankId()].y, off_y);
            atomicMax(&this->data.corner_lower[__xThreadInfo.BankId()].x, off_x);
            atomicMax(&this->data.corner_lower[__xThreadInfo.BankId()].y, off_y);
        }
    }
    if(__xThreadInfo.WarpId()==mBankSize-1){
        for(int i=0;i<offset;i++){
            if(this->data.grid.data.multi_d[off_x][off_y]
                [this->data.flags[__xThreadInfo.BankId()].get_ucRed()]
                [__xThreadInfo.BankId()].get_xCell()!=mEMPTY_CELL){
                    atomicMin(&this->data.corner_upper[__xThreadInfo.BankId()].x, off_x);
                    atomicMin(&this->data.corner_upper[__xThreadInfo.BankId()].y, off_y);
                    atomicMax(&this->data.corner_lower[__xThreadInfo.BankId()].x, off_x);
                    atomicMax(&this->data.corner_lower[__xThreadInfo.BankId()].y, off_y);
        }
    }
}
\end{minted}

\begin{landscape}
\begin{table}[h]
    \centering
    \includegraphics[width=1.0\linewidth]{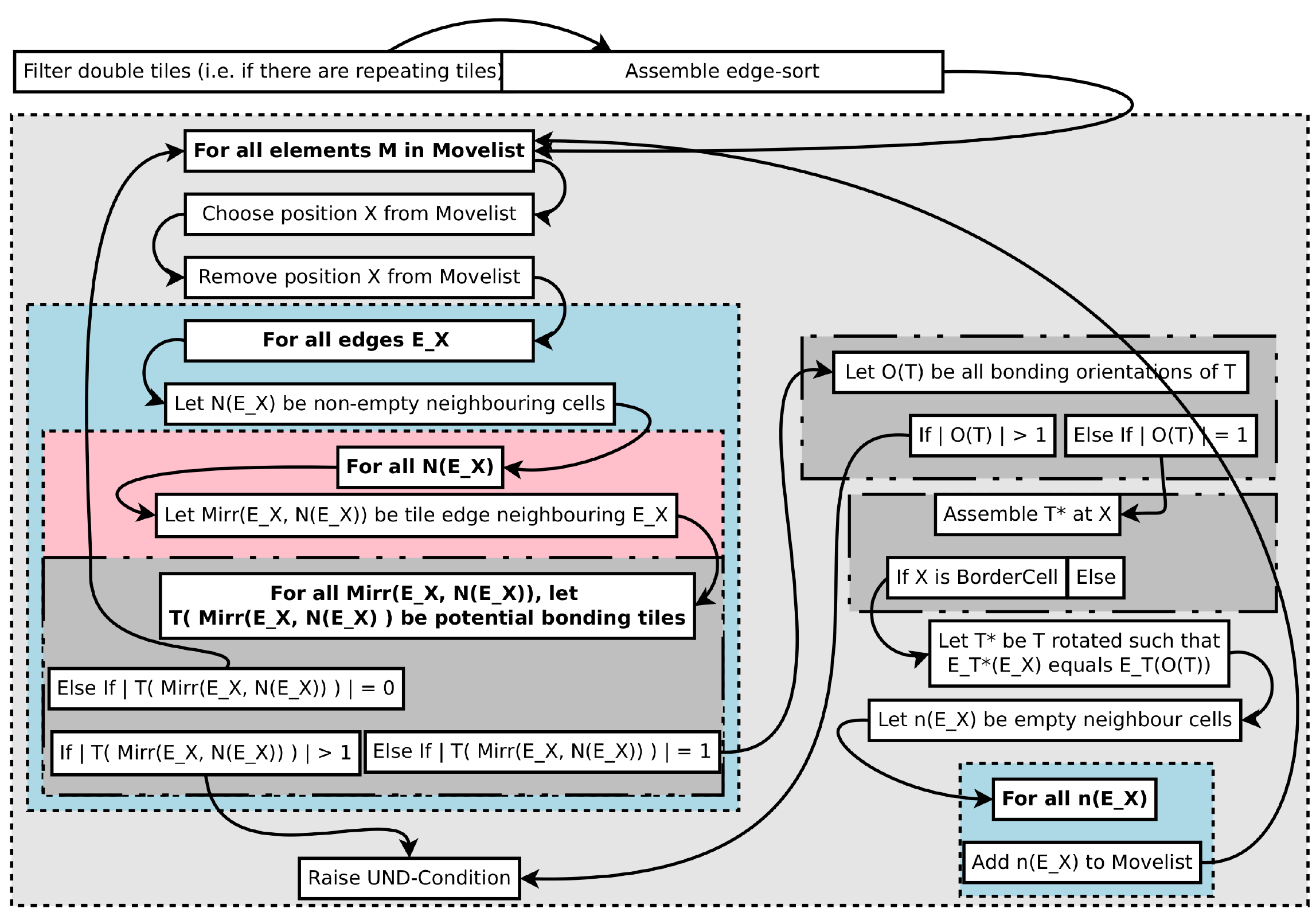}
    \caption{A simplified set-theory inspired flow chart for the movelist polyomino assembly algorithm used in this paper.}
    \label{app:flowchart}
\end{table}
\pagebreak
\begin{table}[h]
    \centering
    \includegraphics[width=1.0\linewidth]{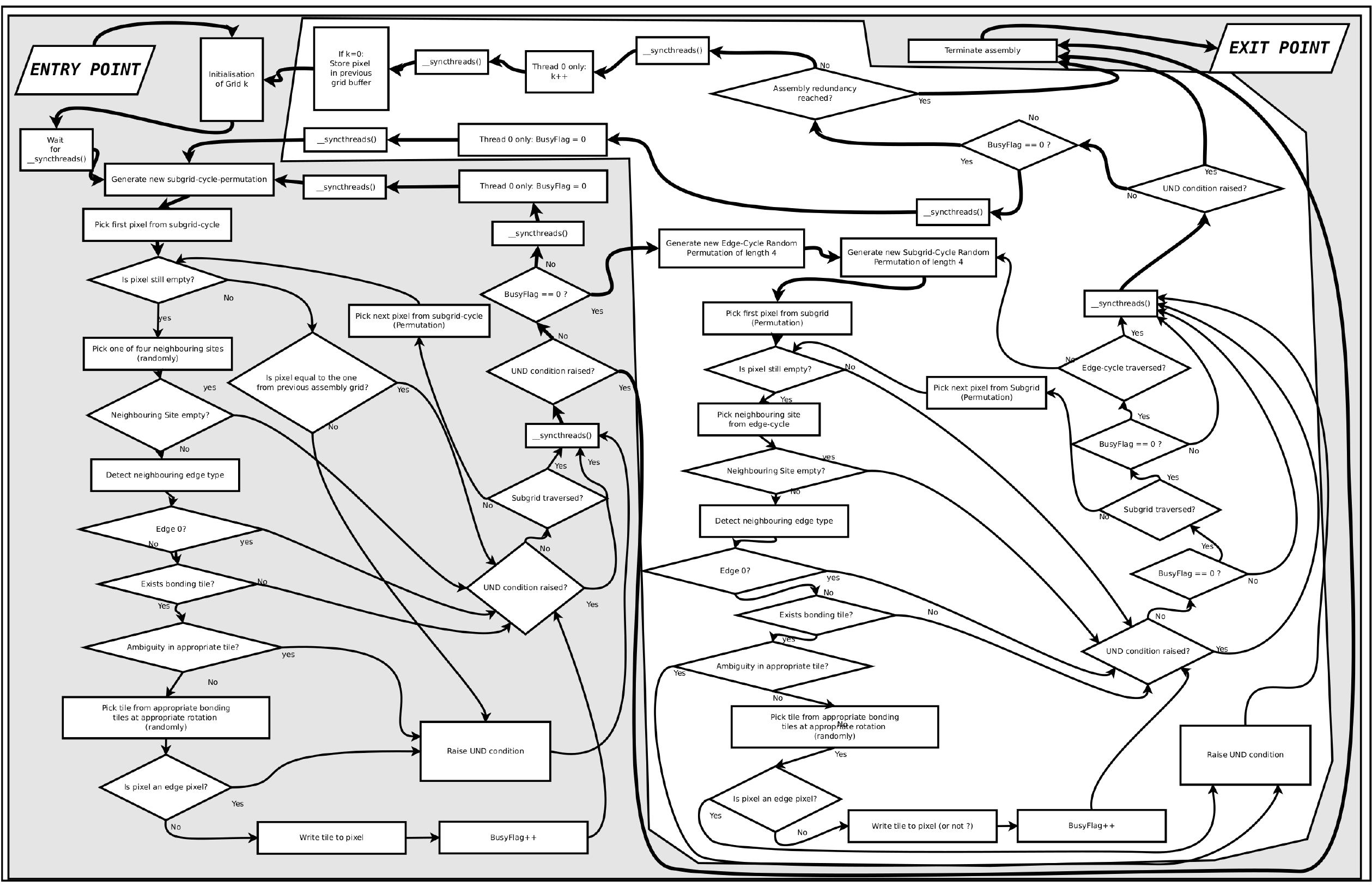}
    \caption{An optimised flow chart for an in-place polyomino assembly algorithms which could be used to take over in case that movelist storage does not suffice.}
    \label{app:movelist}
\end{table}
\end{landscape}

\section{Determination of the maximally required movelist size}

\includegraphics{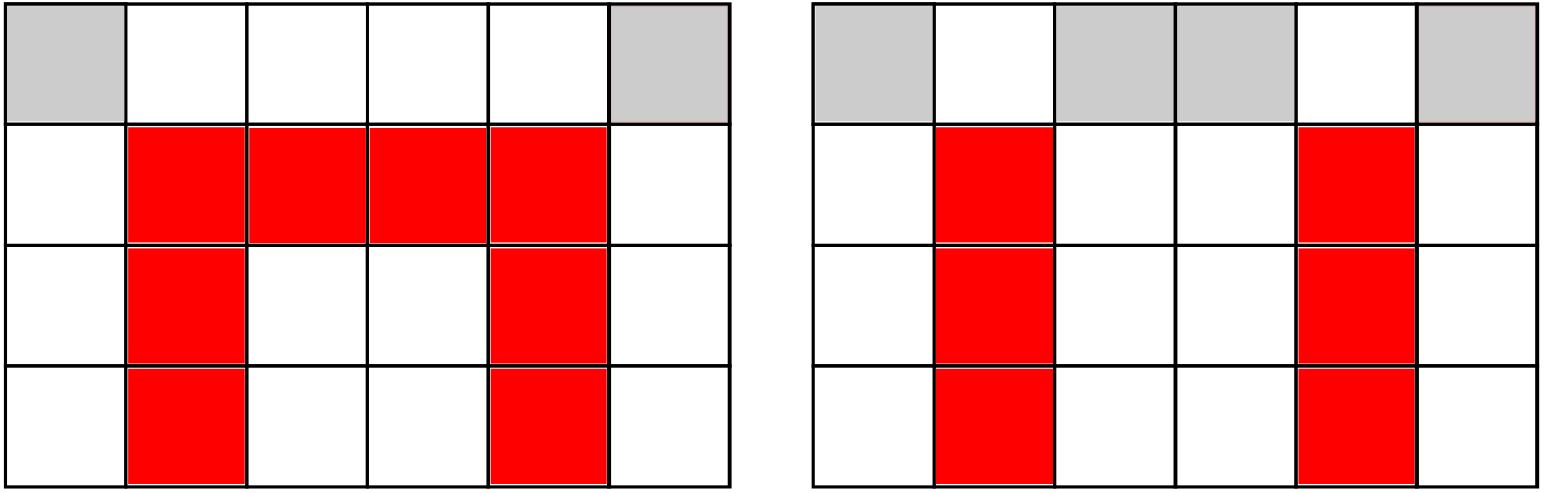}

Joining two open ends is movelist-size-wise equivalent to leaving them unjoined.

\includegraphics{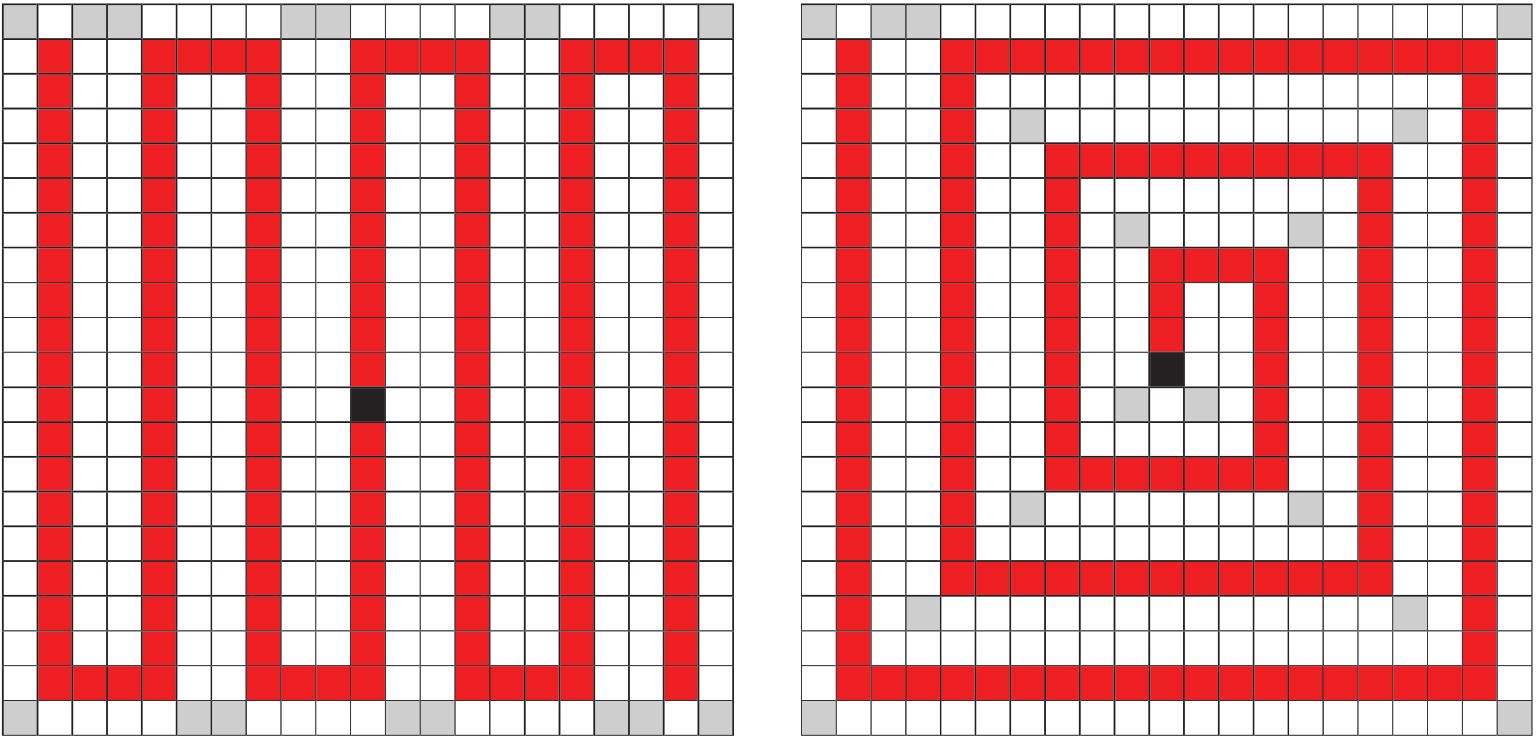}

Left: A growth pattern suspected to have maximum movelist occupation for \textit{LIFO}-stacks.
Right: A growth pattern with maximum movelist occupation for \textit{dynamic} lists. The black tile
is the \textit{seed} tile. Note that the spiral on the right has $13$ kinks, while the structure on the left
has one less - this explains why the spiral is expected to have a smaller maximum movelist than then the left-hand side.

\end{document}